\documentclass[letterpaper, 10 pt, conference]{ieeeconf}  

\usepackage{amsmath,amssymb,amsfonts,amsthm,amscd,dsfont}
\usepackage{algorithm,algorithmicx, listings}
\usepackage[noend]{algpseudocode}
\usepackage{booktabs}
\usepackage{adjustbox,graphicx,subcaption,tabularx}
\usepackage{multicol,multirow}

\usepackage{array}      
\usepackage{rotating}   
\usepackage{cite}
\usepackage{xcolor}

\usepackage{makecell}
\usepackage{multirow}
\usepackage{booktabs}
\usepackage[breaklinks=true, colorlinks, bookmarks=true]{hyperref}  

\IEEEoverridecommandlockouts    
\newtheorem{example}{Example}
\newtheorem{problem}{Problem}

\DeclareMathOperator*{\argmin}{argmin}

\title{\LARGE \bf Closed-Loop Object-Informed Control for Non-Prehensile Robot Manipulation}

\author{Nikola Raicevic, Hyomuk Kim$^{*}$, Shahid Mulla$^{*}$, Hyungjun Doh, Bharath Raam Radhakrishnan, Chenbin Yu, \\ Ki Myung Brian Lee, and Nikolay Atanasov
\thanks{$^{*}$These authors contributed equally to this work.}
\thanks{All authors are with the Department of Electrical and Computer Engineering, University of California San Diego, 9500 Gilman Drive, La Jolla, CA 92093, United States.}
\thanks{This work was supported by the Ministry of Trade, Industry, and Energy (MOTIE), Korea, under the Strategic Technology Development Program supervised by the Korea Institute for Advancement of Technology (KIAT) [Grant No. P0026052].}
}

\begin{document}
\bstctlcite{BSTcontrol}
\maketitle
\thispagestyle{empty}
\pagestyle{empty}
    
\begin{abstract}
\label{sec:abstract}
    Non-prehensile robot manipulation is challenging due to discontinuous, long-horizon interactions between the robot and the objects it manipulates. Sampling-based model predictive control methods are effective with discontinuous contact but face challenges with finding promising trajectories in long-horizon planning. We propose a closed-loop object-informed (CLOI) method that splits the problem into object-level planning to find long-horizon object poses that lead the object to its goal, and robot-level planning to select robot actions that follow those poses. We use model predictive path integral (MPPI) control to solve the subproblems and couple their solutions through consensus on the object poses using the alternating direction method of multipliers (ADMM). The object plan is revised toward robot-realizable object trajectories, while the robot plan is aligned with the object poses the task requires. In planar pushing tasks with obstacles using an xArm6 manipulator, CLOI increases the success rate by 35\% in simulation and 43\% on hardware, compared to standard MPPI given the same computational budget. Project github: \href{https://github.com/ExistentialRobotics/CLOI.git}{CLOI.github.io}.
\end{abstract}

\section{Introduction}\label{sec:introduction}

\begin{figure}[t]
    \centering
    \includegraphics[width=\linewidth,clip]{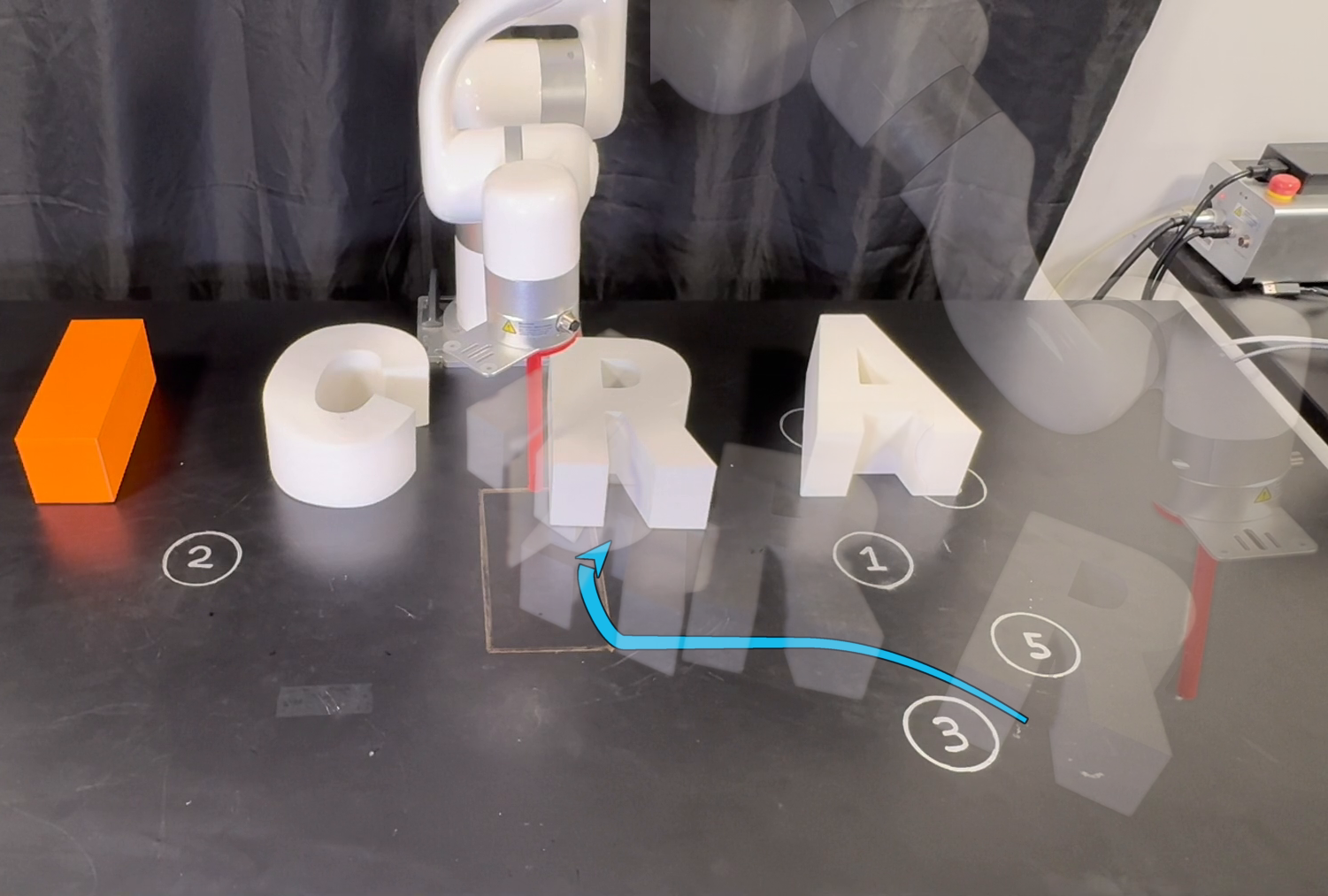}
    \caption{An xArm6 pushes the letter \emph{R} to complete the word ``ICRA'' while avoiding \emph{I},\emph{C},and \emph{A}, using the proposed \emph{CLOI} approach.}
    \label{fig:experiment_setup}
\end{figure}

Non-prehensile manipulation allows a robot to move objects that are too large, heavy, or flat to grasp, and to rearrange them in clutter where no collision-free grasp exists (Fig.~\ref{fig:experiment_setup}). Recent advances in modeling and simulation of multi-body contact dynamics \cite{makoviychuk2021isaac, todorov2012mujoco} have enabled object motion prediction under intermittent contact in parallel and at faster-than-real-time speeds. Sampling-based model predictive control (MPC), and in particular model predictive path integral (MPPI) control \cite{williams2017informationtheoreticmodelpredictive}, can utilize such fast parallelized simulators to achieve real-time control given a suitable cost description of the manipulation task~\cite{pezzato2025samplingbasedmodelpredictivecontrol}.

However, long-horizon planning for non-prehensile manipulation remains challenging for two reasons. First, contact dynamics are discontinuous, so small changes in robot motion can produce substantially different object behaviors depending on how contact is established or broken. This renders large portions of the robot's configuration space not useful for task progress. Second, the object is actuated only implicitly, and the robot must select its own motion to produce the desired object displacement through contact. This dramatically expands the search space, as the planner must simultaneously reason about robot kinematics, contact geometry, and object response.

\begin{figure*}[t]
    \centering
    \includegraphics[width=0.95\linewidth]{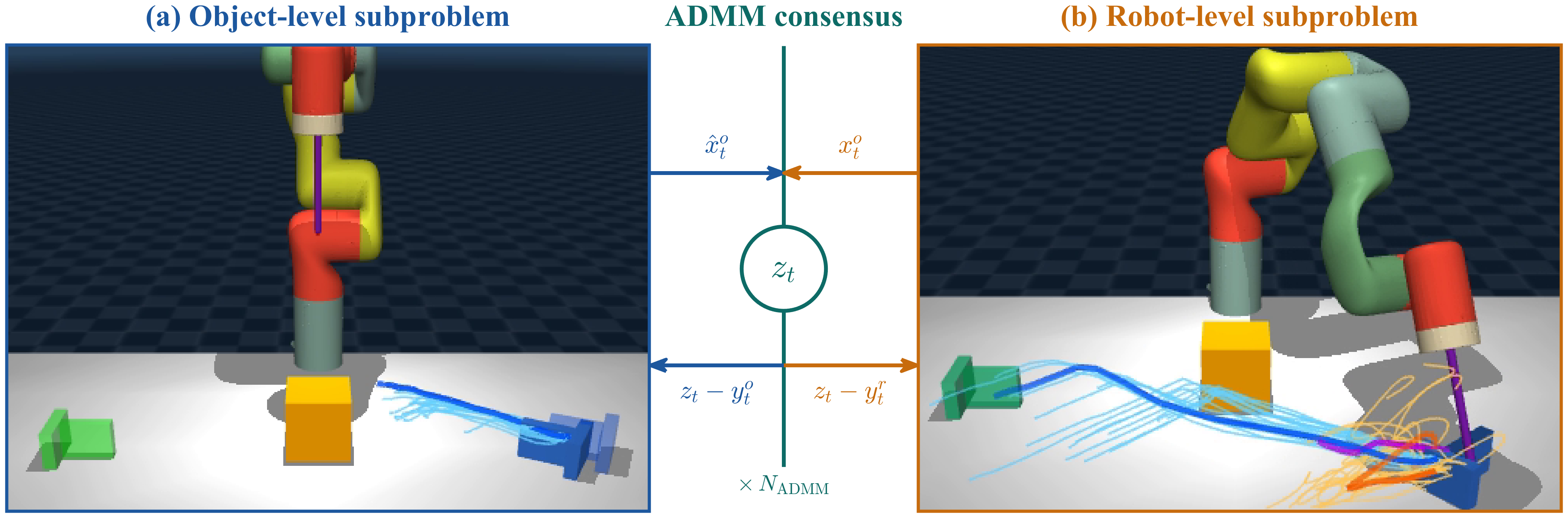}
    \caption{\textbf{Closed-loop object-informed non-prehensile manipulation:} (a) An object-level subproblem generates object trajectories (blue) past obstacles (yellow) to a desired goal (green); (b) A robot-level subproblem generates robot end-effector trajectories (orange) that induce different object trajectories (magenta). Light lines are samples and bold lines indicate the optima. ADMM is used to achieve agreement on the object poses between the object-level and the robot-level planners. The desired object poses of each block are averaged in a consensus variable $z_t$ and used to update the dual variables $y_t^i$ and the consensus target $z_t - y_t^i$ in the rollout cost.}
    \label{fig:pictorial_diagram}
\end{figure*}

To address these challenges, we decompose the joint object-robot problem into two subproblems coupled through the wrench that produces the object pose, which we use as a consensus variable, as illustrated in Fig.~\ref{fig:pictorial_diagram}. 
The object-level subproblem plans an object trajectory to the goal assuming it is directly actuated, as shown in Fig.~\ref{fig:pictorial_diagram}(a), while the robot-level subproblem generates robot actions that achieve the desired object trajectory, as illustrated in Fig.~\ref{fig:pictorial_diagram}(b). The object-level planner operates in a lower-dimensional space, making it easier to converge and plan an object's trajectory with greater foresight. The robot-level planner operates in a high-dimensional space and is able to adjust the target object well locally, but lacks the long-term planning ability. Two subproblems are coordinated by the alternating direction method of multipliers (ADMM)~\cite{boyd2011distributedoptimizationandstatistical} using the object pose as a consensus variable to achieve agreement on the desired object trajectory. The process is repeated in a closed loop, where the object plan is revised towards trajectories that the robot can realize, while the robot's actions are steered towards object trajectories that achieve the task. This motivates the name of our method: closed-loop object-informed (CLOI) control. Our contributions are summarized as follows.
\begin{itemize}
    \item We formulate non-prehensile manipulation as a two-block consensus ADMM problem with object-level and robot-level subproblems that agree on the desired sequence of object poses, and solve these subproblems using sampling-based model predictive control (MPC). 
    
    \item In planar pushing tasks with obstacles using an xArm6 manipulator, CLOI increases the success rate by \textbf{35\%} in simulation and \textbf{43\%} on hardware, compared to joint MPPI planning with a matched computational budget.
\end{itemize}

\section{Related Work}
\label{sec:related_work}

Non-prehensile manipulation is challenging because of discontinuous contact dynamics, prompting various formulations of contact and dynamics models. Contact forces can be modeled with complementarity constraints \cite{trinkle1996implicit} or an equivalent second-order cone program (SOCP) \cite{anitescu2006optimization}. Recent work optimizes through such analytical models by locally approximating the complementarity conditions \cite{posa2014adirectmethodfortrajectoryoptimization,li2025surprising,graesdal2024tightconvexrelaxationscontactrich}, smoothing the SOCP~\cite{pang2022localsmoothing}, or constructing an explicit solution~\cite{jin2025complementarityfreemulticontactmodelingoptimization}. These formulations require specialized differentiable contact models, which are difficult to obtain for arbitrary objects.

Sampling-based MPC evaluates control sequences by parallel simulation and requires no gradients. A distribution over control sequences is updated iteratively from the cost of the resulting rollouts \cite{wagener2019online}. Representative methods include the cross-entropy method (CEM)~\cite{deboar2005cem}, MPPI~\cite{williams2017informationtheoreticmodelpredictive}, and diffusion-inspired annealing~\cite{xue2024fullordersamplingbasedmpctorquelevel}, which are effective for manipulation~\cite{bhardwaj2021stormintegratedframeworkfast} on GPU-accelerated simulators~\cite{pezzato2025samplingbasedmodelpredictivecontrol}. Applied monolithically to non-prehensile manipulation, however, these methods must search the joint robot--object space, in which the fraction of samples that lead to object actuation is small. 
Model-based controllers for planar pushing exploit the pusher--slider model and an explicit choice of contact mode (e.g. sticking, sliding, and separation) by replacing hybrid dynamics with a smooth program to improve both performance and computational efficiency~\cite{hogan2016feedbackcontrolpusherslidersystem,federico2025nonlinearmodelpredictivecontrol,lee2015hierarchical,bui2025pushanythingsinglemultiobject}. Closest to our formulation are object-centric planning methods~\cite{ren2025objectcentric,king2016rearrangementplanningusingobjectcentric}, which first plan an object trajectory that achieves the goal and then solve for the robot motion that realizes it. These approaches are unidirectional: the object trajectory is committed before the robot's reachability and frictional limits are known, and the robot cannot revise it.

ADMM~\cite{boyd2011distributedoptimizationandstatistical} splits a coupled problem into blocks that are optimized separately and reconciled through a shared variable. It converges under mild conditions for a family of nonconvex problems~\cite{hong2015convergence}. Consensus complementarity control (C3) applies ADMM to contact-implicit MPC~\cite{venkatesh2025approximatingglobalcontactimplicitmpc}, splitting it into a step that ignores the complementarity constraints and one that enforces them, thereby isolating the non-convexity and exploiting the temporal independence of complementarity constraints. We also use ADMM for splitting, but from a physical intuition: our two blocks are the object and the robot, which are physically distinct subsystems coupled only by the wrench. The object can follow a simpler quasi-static model~\cite{howe1996practical,zhou2017faststochasticcontactmodel}. The split therefore reduces the search for each sampler by providing guidance rather than isolating constraints. Both blocks remain sampling-based, and we adopt stochastic~\cite{ouyang2013stochastic} and efficient~\cite{franca2016explicit,goldstein2014fast} ADMM variants to stabilize the consensus against non-smooth contact.

\section{Problem Statement}
\label{sec:problem_statement}

We consider planar object pushing, where a robot manipulator must push a rigid object across a horizontal support surface to a goal pose. The manipulator is modeled in discrete-time input-affine form,
\begin{equation} \label{eq:robot_dyn_ps}
    x_{t+1}^r \;=\; f_r(x_t^r) \;+\; g_r(x_t^r)\,u_t^r \;+\; h_r(x_t^r)\,w_t^r ,
\end{equation}
where $x_t^r \in \mathbb{R}^{n^r}$ is the robot's configuration (e.g., joint angles), $u_t^r \in \mathbb{R}^{m^r}$ is the control input (e.g., joint velocities), and $w_t^r \in \mathbb{R}^{p^r}$ is the external wrench on the robot induced by contact with the object or the environment.

The object's motion arises through contact with the robot and the environment, modeled as a discrete-time input-affine system,
\begin{equation} \label{eq:object_dyn_ps}
    x_{t+1}^o \;=\; f_o(x_t^o) \;+\; h_o(x_t^o)\,w_t^o ,
\end{equation}
where $x_t^o \in \mathbb{R}^{n^o}$ is the object's pose ($n^o = 3$ for position and orientation in 2D) and $w_t^o \in \mathbb{R}^{p^o}$ is the external wrench induced by contact with the robot or the environment.

The coupling between the robot and the object arises through the external wrench terms $w_t^r$ and $w_t^o$. We consider a frictionless model of contact:
\begin{equation}
\label{eq:wrench_coupling}
\begin{aligned}
    w_t^o &= \begin{bmatrix} F_t^o \\ \tau_t^o \end{bmatrix} = \sum_{j \in \{1,2\}} \lambda_t^j
        \begin{bmatrix} \nabla_{p^o} \phi_j \\
        (p_t^{c,j} - p_t^o) \times \nabla_{p^o} \phi_j \end{bmatrix},\\
    w_t^r &= \begin{bmatrix} F_t^r \\ \tau_t^r \end{bmatrix} = \sum_{j \in \{1,3\}} \lambda_t^j
        \begin{bmatrix} \nabla_{p^r} \phi_j \\
        (p_t^{c,j} - p_t^r) \times \nabla_{p^r} \phi_j \end{bmatrix},
\end{aligned}
\end{equation}
where $\phi_j$ is a signed distance function encoding the distance between two rigid bodies: (i) $\phi_1$ between the robot and the object, (ii) $\phi_2$ between the object and the environment, and (iii) $\phi_3$ between the robot and the environment, each defined so that $\phi_j \ge 0$ corresponds to non-penetration and $\phi_j = 0$ to contact at the boundary. The scalar $\lambda_t^j \in \mathbb{R}_{\ge 0}$ is the normal contact force magnitude associated with $\phi_j$,\footnote{For multiple simultaneous contacts, $\phi_j$ and $\lambda_t^j$ become vector-valued} $p_t^{c,j}$ is the contact point of $\phi_j$, and $p_t^o$, $p_t^r$ are the object and robot position components about which the torques are taken. Since each $\phi_j$ is a signed distance, its position gradient $\nabla_{p^i} \phi_j$ is the unit contact normal at $p_t^{c,j}$ directed into body $i \in \{o, r\}$, so $\lambda_t^j \nabla_{p^i} \phi_j$ is the force that contact applies to the body; summing over contacts gives the net force $F_t^i \in \mathbb{R}^2$ and the net torque $\tau_t^i \in \mathbb{R}$ about $p_t^i$. Contact $1$ is shared and $\nabla_{p^o} \phi_1 = -\nabla_{p^r} \phi_1$, so the object and the robot receive equal and opposite forces from the single magnitude $\lambda_t^1$. The magnitudes are subject to complementarity constraints~\cite{trinkle1996implicit},
\begin{equation}
\label{eq:complementarity}
\begin{aligned}
    0 \le \lambda_t^1 &\perp \phi_1(x_t^o, x_t^r) \ge 0, \\
    0 \le \lambda_t^2 &\perp \phi_2(x_t^o) \ge 0, \\
    0 \le \lambda_t^3 &\perp \phi_3(x_t^r) \ge 0 ,
\end{aligned}
\end{equation}
where $a \perp b$ denotes $a^\top b = 0$. This encodes that contact forces are non-zero only when the corresponding bodies are in contact ($\phi_j = 0$), and that bodies cannot interpenetrate ($\phi_j \ge 0$). Other environmental obstacles are captured by $\phi_2$ and $\phi_3$ and are assumed to remain static throughout.

We rely on access to a physics simulator that implements
\eqref{eq:robot_dyn_ps}--\eqref{eq:complementarity}, e.g., NVIDIA Isaac~\cite{makoviychuk2021isaac} or MuJoCo~\cite{todorov2012mujoco}. The following example illustrates that the robot and object dynamics can be written in the form we assume.

\begin{example}[Quasi-static planar pushing]
\label{ex:qs_pushing}
Consider a point robot at $x_t^r \triangleq [p_{x,t}^{r},\, p_{y,t}^{r}]^\top \in \mathbb{R}^2$ and a planar object $\mathcal{B} \subset \mathbb{R}^2$ with signed distance function
$d_{\mathcal{B}}(x) = \pm \min_{b \in \partial\mathcal{B}}\|b - x\|$, positive outside $\mathcal{B}$. The object pose $x_t^o \triangleq [p_{x,t}^{o},\, p_{y,t}^{o},\, \theta_t^o]^\top \in \mathbb{R}^3$ contains its position and orientation, and its external wrench is $w_t^o \triangleq [F_{x,t}^{o},\, F_{y,t}^{o},\, \tau_t^o]^\top \in \mathbb{R}^3$. Under quasi-static mechanics, the body twist is proportional to the applied wrench through the ellipsoidal limit surface~\cite{howe1996practical,zhou2017faststochasticcontactmodel},
\begin{equation}
\label{eq:qs_limit_surface}
    \dot x_t^o = D\,w_t^o, \qquad
    D = \operatorname{diag}\big(\mu m g,\, \mu m g,\, c\,r\,\mu m g\big)^{-1},
\end{equation}
with $\mu$ the support friction coefficient, $m$ the object mass, $g$ gravity, $r$ a characteristic length, and $c > 0$ the limit-surface shape parameter. The support contact is not modeled explicitly, as $D$ encodes its frictional resistance. Euler discretization of \eqref{eq:qs_limit_surface} yields the object dynamics:
\begin{equation}
\label{eq:qs_object_dt}
    x_{t+1}^o
    \;=\;
    \underbrace{x_t^o}_{f_o(x_t^o)}
    \;+\;
    \underbrace{\Delta t\,D}_{h_o(x_t^o)}\,w_t^o .
\end{equation}

To show that the wrench follows the form in \eqref{eq:wrench_coupling}, we express $\phi_1$ through $d_{\mathcal{B}}$,
\begin{equation}
\label{eq:qs_gap}
    \phi_1(x_t^o, x_t^r) = d_{\mathcal{B}}\!\big(R^\top(\theta_t^o)(x_t^r - p_t^o)\big),
\end{equation}
with $R(\theta)$ denoting a 2D rotation matrix. A point robot contacts the object at its own position, so $p_t^{c,1} = x_t^r$ and
\begin{equation*}
    w_t^o \;=\; \lambda_t^1\, \nabla_{x^o} \phi_1(x_t^o, x_t^r)
    \;=\; \lambda_t^1
    \begin{bmatrix}
        -R(\theta_t^o)\,\nabla d_{\mathcal{B}} \\
        \nabla d_{\mathcal{B}} \times R^\top(\theta_t^o)(x_t^r - p_t^o)
    \end{bmatrix},
\end{equation*}
with $\nabla d_{\mathcal{B}}$ evaluated at $R^\top(\theta_t^o)(x_t^r - p_t^o)$. Without environment interaction, this is the only contact, so the object's wrench is equal and opposite to the robot's wrench.
\end{example}

\begin{problem}\label{prob:main}
Given initial configurations $(x_0^r, x_0^o)$ and an object goal $g \in \mathbb{R}^{n^o}$, we seek a sequence of robot controls
$\mathbf{U}^r \triangleq \{u_t^r\}_{t=0}^{H-1}$ over a horizon $H \in \mathbb{N}$ that solves
\begin{align}
    \min_{\mathbf{U}^r} \quad
    & \sum_{t=0}^{H-1} \Big( \ell_o(x_t^o) + \ell_r(x_t^r, u_t^r) \Big)
      + \ell_f(x_H^o) \notag\\
    \text{s.t.}\quad
    & \text{dynamics and contact }
      \eqref{eq:robot_dyn_ps}\text{--}\eqref{eq:complementarity},  \label{eq:problem_formulation}\\
    & (x_t^r, u_t^r) \in \mathcal{C}, \quad t = 0,\dots,H-1 . \notag
\end{align}
Here, $\mathcal{C} \subset \mathbb{R}^{n^r} \times \mathbb{R}^{m^r}$ encodes constraints such as velocity and actuation limits. The functions $\ell_o$, $\ell_f$, and $\ell_r$ are the object stage cost, object terminal cost, and robot stage cost, respectively, which encode the pushing task, e.g., by measuring the distance from the object state $x_t^o$ to the goal $g$ and regularizing the magnitude of the robot control inputs. The object is unactuated, so $\mathbf{U}^r$ is the only decision variable, and it affects $\ell_o$ and $\ell_f$ through the contact dynamics \eqref{eq:object_dyn_ps}--\eqref{eq:complementarity}, whose complementarity structure makes \eqref{eq:problem_formulation} non-convex.
\end{problem}

\section{Closed-Loop Object-Informed Control}
\label{sec:technical_approach}

Problem~\ref{prob:main} is challenging because the object moves through intermittent contact produced by the robot. A high-dimensional robot arm has a large control space in which most actions fail to contact the object. Hence, sampling-based planning methods for \eqref{eq:problem_formulation} are inefficient without guidance, as only a few rollouts move the object.

Our key idea is to guide the search by decoupling the joint object-robot problem \eqref{eq:problem_formulation} into 1) an object-level subproblem that finds a wrench sequence to drive the object to its goal, as if the object could self-actuate; and 2) a robot-level subproblem that plans robot actions guided by the object plan. The object-level subproblem is lower-dimensional and much simpler compared to the full problem in \eqref{eq:problem_formulation} and, hence, easier to converge, leading to a more efficient plan. To ensure that the two subproblems agree on the object motion, we use the ADMM algorithm, which steers the object plan towards object poses that the robot can realize and the robot plan towards object poses that the task demands. 

\subsection{Object-Level and Robot-Level Subproblems}
\label{subsec:subproblems}
We first decouple \eqref{eq:problem_formulation} into object-level and robot-level subproblems. Intuitively, we consider an object that is able to self-actuate by controlling the wrench $w_t^o$ in \eqref{eq:object_dyn_ps} directly. Taking the wrench sequence $\mathbf{U}^o \triangleq \{{w}_t^o\}_{t=0}^{H-1}$ over a planning horizon $H$ as the decision variable and dropping all robot-related variables from \eqref{eq:problem_formulation}, gives an object-level subproblem:
\begin{align}
\label{eq:object_subproblem}
    \min_{\mathbf{U}^o\in\mathcal{C}^o} \quad
    & J_o(\mathbf{U}^o) \;\triangleq\; \sum_{t=0}^{H-1} \ell_o({x}_t^o, {w}_t^o) \;+\; \ell_f({x}_{H}^o) \\
    \text{s.t.}\quad
        & \;\text{dynamics and contact } \eqref{eq:object_dyn_ps}\text{--}\eqref{eq:complementarity}, \notag
\end{align}
where $\mathbf{U}^o \in \mathcal{C}^o$ represents friction cone constraints. The solution to \eqref{eq:object_subproblem} is an object wrench sequence $\mathbf{U}^{o^*}$ and associated object pose trajectory ${x}_{t:t+H}^{o^*}$, which can be used to guide the robot-level subproblem. In other words, we can define a robot-level subproblem that aims to track the sequence ${x}_{t:t+H}^{o^*}$ determined in the object-level subproblem:
\begin{align}
\label{eq:robot_subproblem}
    \min_{\mathbf{U}^r \in \mathcal{C}^r} \;
    & \sum_{t=0}^{H-1} \!\Big( \ell_r(x_t^r, u_t^r) + \frac{\rho}{2} \|x_t^o - {x}_t^{o^*}\|^2  \Big)\!
      + \frac{\rho}{2} \|x_H^o - {x}_H^{o^*}\|^2 \notag\\
    \text{s.t.}\;
    & \;\text{dynamics and contact }
      \eqref{eq:robot_dyn_ps}\text{--}\eqref{eq:complementarity},
\end{align}
where $\rho$ is the weight of the object trajectory tracking term $\|x_t^o - {x}_t^{o^*}\|^2$.  This term provides long-horizon guidance to the robot from the object-level subproblem. The constraint $\mathcal{C}^r$ is as per~\eqref{eq:problem_formulation}.
For conciseness, we will abbreviate the robot-level cost by $J_r(\mathbf{U}^r) \triangleq \sum_{t=0}^{H-1} \ell_r(x_t^r, u_t^r)$.

The intuition described above can be formalized through a consensus ADMM formulation \cite[Ch.~7]{boyd2011distributedoptimizationandstatistical}. Doing so introduces a bidirectional consensus, rather than the unidirectional hierarchy in~\eqref{eq:robot_subproblem}. We introduce a consensus variable $z_t$, whose role is to achieve agreement on the object poses ${x}_t^o$ optimized in the object-level subproblem and the object poses $\hat x_t^o$ optimized in the robot-level subproblem through contact between the robot and the object. Starting with the joint problem in \eqref{eq:problem_formulation}, we introduce the consensus variable $z_t$:
\begin{equation}
\label{eq:consensus_problem}
\begin{aligned}
    \min_{\mathbf{U}^o \in \mathcal{C}^o,\, \mathbf{U}^r \in \mathcal{C}^r,\, \mathbf{Z}} \quad
    & J_o(\mathbf{U}^o) + J_r(\mathbf{U}^r) \\
    \text{s.t.}\quad
    & A_t^o(\mathbf{U}^o) = {x}_t^o = z_t, \;\; t = 0,\dots,H-1,    \\
    & A_t^r(\mathbf{U}^r) = \hat x_t^o = z_t, \;\; t = 0,\dots,H-1, \\
\end{aligned}
\end{equation}
where $\mathbf{Z} \triangleq \{z_t\}_{t=0}^{H-1}$ are the consensus variables. The notation $A_t^o(\mathbf{U}^o)$ refers to the computation of the object state ${x}_t^o$ by propagating the wrenches $\mathbf{U}^o$ through the object dynamics in \eqref{eq:object_dyn_ps}--\eqref{eq:complementarity}. Similarly, $A_t^r(\mathbf{U}^r)$ refers to the computation of the object state $\hat x_t^o$ by propagating the robot inputs $\mathbf{U}^r$ through the coupled robot-object dynamics in \eqref{eq:robot_dyn_ps}--\eqref{eq:complementarity}.

ADMM forms the augmented Lagrangian of \eqref{eq:consensus_problem} with dual variables $y_t^o$, $y_t^r$ and penalty $\rho>0$:
\begin{equation} \label{eq:augmented_lagrangian}
    \mathcal{L} = \sum_{i \in \{o,r\}}\Bigl( J_i(\mathbf{U}^i)
    + \frac{\rho}{2} \sum_{t=0}^{H-1}
      \| A_t^i(\mathbf{U}^i) - z_t + y_t^i \|^2 \Bigr),
\end{equation}
where $y_t^i$ are the dual variables for the consensus constraints in~\eqref{eq:consensus_problem}.
Unlike~\eqref{eq:object_subproblem}, the augmented Lagrangian also steers the object problem toward the robot problem.
ADMM minimizes it by alternating between a primal update for each block (object- and robot-levels), a consensus variable update for $\mathbf{Z}$, and dual variable updates as described in Algorithm~\ref{alg:cloi}. 

\begin{algorithm}[t]
\caption{\small Closed-Loop Object-Informed Control}
\label{alg:cloi}
\small
\begin{algorithmic}[1]
\Require State $x_t=(x_t^r,x_t^o)$; horizon $H$; samples $K^o,K^r$; previous $(\mathbf{U}^{o},\mathbf{U}^{r},\mathbf{Z},\mathbf{Y}^o,\mathbf{Y}^r)$; parameters $\rho,\gamma,\nu$
\State Warm-start $(\mathbf{U}^{o,(0)},\mathbf{U}^{r,(0)},\mathbf{Z}^{(0)},\mathbf{Y}^{o,(0)},\mathbf{Y}^{r,(0)})$ by shifting the previous solution
\For{$l=0,\dots,N_{\mathrm{ADMM}}-1$}
    \State \textbf{Object update:} solve \eqref{eq:block_update_cloi} for $i=o$ with Alg.~\ref{alg:mppi}
    \State \textbf{Robot update:} solve \eqref{eq:block_update_cloi} for $i=r$ with Alg.~\ref{alg:mppi}
    \State \textbf{Consensus update:} $z_t^{(l+1)}$ by \eqref{eq:consensus_update_cloi}, $\forall t$
    \State \textbf{Dual update:} $y_t^{i,(l+1)}$ by \eqref{eq:dual_update_cloi}, $\forall t$, $i \in \{o,r\}$
    \State \textbf{if} $\|r^{(l+1)}\| \le \epsilon_r$ and $\|d^{(l+1)}\| \le \epsilon_s$ \textbf{then break}
\EndFor
\State Apply $u_t^r \gets \bar u_0^r$; observe $x_{t+1}$; set $t \gets t+1$
\end{algorithmic}
\end{algorithm}

\noindent\textbf{Primal update.} Each block minimizes its own subproblem within \eqref{eq:augmented_lagrangian} with $z_t$ and $y_t^i$ held fixed,
\begin{align}
    \mathbf{U}^{i,(l+1)} &= \argmin_{\mathbf{U}^i \in \mathcal{C}_i} \; S^{i,(l)}(\mathbf{U}^i),
    \label{eq:block_update_cloi} \\
    \text{where} \quad S^{i,(l)}(\mathbf{U}^i) &\triangleq J_i(\mathbf{U}^i)
        + \frac{\gamma}{2}\bigl\|\mathbf{U}^i - \mathbf{U}^{i,(l)}\bigr\|^2 \notag \\
    &\quad + \frac{\rho}{2}\sum_{t=0}^{H-1}\bigl\|A_t^i(\mathbf{U}^i) - z_t^{(l)} + y_t^{i,(l)}\bigr\|^2 . \notag
\end{align}
The proximal and penalty terms, weighted by $\gamma$ and $\rho$, penalize disagreement between two subproblems.

\noindent\textbf{Consensus update.} Minimizing \eqref{eq:augmented_lagrangian} in $\mathbf{Z}$ averages the consensus variables and their duals, which represent accumulated disagreement. We weight that average,
\begin{align}
    z_t^{(l+1)} &= \nu\big(A_t^o(\mathbf{U}^{o,(l+1)}) + y_t^{o,(l)}\big) \notag\\
    &\quad + (1-\nu)\big(A_t^r(\mathbf{U}^{r,(l+1)}) + y_t^{r,(l)}\big). \label{eq:consensus_update_cloi}
\end{align}
so that $\nu \in (0,1)$ sets how far the agreement leans toward each block. Larger $\nu$ favors the object plan and smaller $\nu$ favors the robot plan, so $\nu = 1$ ignores the robot plan entirely, and vice versa for $\nu = 0$.

\noindent\textbf{Dual update.} The dual variables accumulate the remaining primal residual, so that persistent disagreement is penalized more heavily at the next iteration,
\begin{equation}
\label{eq:dual_update_cloi}
    y_t^{i,(l+1)} = y_t^{i,(l)} + A_t^i(\mathbf{U}^{i,(l+1)}) - z_t^{(l+1)},
    \quad i \in \{o, r\} .
\end{equation}
The termination criteria thresholds primal and dual residuals:
\begin{equation}
\label{eq:residuals}
    r^{i,(l+1)} = A^i(\mathbf{U}^{i,(l+1)}) - \mathbf{Z}^{(l+1)},
    \; \;d^{(l+1)} = \rho\big(\mathbf{Z}^{(l+1)} - \mathbf{Z}^{(l)}\big).
\end{equation}

\subsection{Sampling-Based Solution of the Subproblems}
\label{subsec:mppi_foundations}

\begin{algorithm}[t]
\caption{\small Standard MPPI Iteration}
\label{alg:mppi}
\small
\begin{algorithmic}[1]
    \Require State $x_t$, nominal inputs $\bar{\mathbf{U}}_t^r$,
    exploration covariance $\Sigma_u$
    \For{$k=1,\dots,K$}
        \For{$h=0,\dots,H-1$}
            \State Sample $\epsilon_{t+h}^{(k)} \sim \mathcal{N}(0,\Sigma_u)$
            \State $u_{t+h}^{i,(k)} \gets \bar u_{t+h}^i + \epsilon_{t+h}^{(k)}$
        \EndFor
        \State Roll out dynamics using \eqref{eq:robot_dyn_ps}--\eqref{eq:complementarity}
        \State Evaluate $S^{i,l,(k)}$ using \eqref{eq:block_update_cloi}
    \EndFor
    \State Update $\bar{\mathbf{U}}_t^r$ using \eqref{eq:mppi_reweighting}
    \State \Return $\bar{\mathbf{U}}_t^r$
\end{algorithmic}
\end{algorithm}

The object and robot subproblems in \eqref{eq:block_update_cloi} can be solved by an optimization method, such as MPPI~\cite{williams2017informationtheoreticmodelpredictive}. MPPI is a derivative-free sampling-based method for finite-horizon optimal control problems, which is well suited to contact dynamics because it requires only forward rollouts of \eqref{eq:robot_dyn_ps}--\eqref{eq:complementarity}. MPPI maintains a nominal control sequence $\bar{\mathbf{U}}_t^i \triangleq \{\bar u_{t+h}^i\}_{h=0}^{H-1}$. Each iteration draws $K$ perturbations:
\begin{equation}
\label{eq:mppi_sampling}
    u_{t+h}^{i,(k)} = \bar u_{t+h}^i + \epsilon_{t+h}^{(k)}, \qquad
    \epsilon_{t+h}^{(k)} \sim \mathcal{N}(0,\Sigma_u),
\end{equation}
where $\Sigma_u \in \mathbb{R}^{m^r \times m^r}$ is the exploration covariance. Each perturbed sequence is rolled out using \eqref{eq:robot_dyn_ps}--\eqref{eq:complementarity} and scored by the objective in \eqref{eq:block_update_cloi}, yielding its sample cost $S^{i,l,(k)}$. The nominal sequence is then updated by an importance-weighted average of the perturbations as:
\begin{equation}
\label{eq:mppi_reweighting}
    \bar u_{t+h}^i \leftarrow \bar u_{t+h}^i
    + \sum_{k=1}^{K} \frac{ \epsilon_{t+h}^{(k)}\exp\!\left(-\frac{1}{\lambda}\left(S^{i,l,(k)}-S_{\min}\right)\right)}
    {\sum_{j=1}^{K}\exp\!\left(-\frac{1}{\lambda}\left(S^{i,l,(j)}-S_{\min}\right)\right)}
\end{equation}
where $S_{\min} \triangleq \min_{j} S^{i,l,(j)}$ and $\lambda > 0$ sets the selectivity of the update. MPPI runs in receding horizon, applying $u_t^i = \bar u_t^i$, shifting $\bar{\mathbf{U}}_{t+1}^i \gets \mathrm{shift}(\bar{\mathbf{U}}_t^i)$, and re-optimizing at time $t{+}1$. Algorithm~\ref{alg:mppi} summarizes one iteration.

\section{Experimental Results}
\label{sec:experiments}

\begin{figure*}[thpb]
    \centering
    \begin{minipage}[t]{0.16\linewidth}
        \centering
        \includegraphics[width=\linewidth]{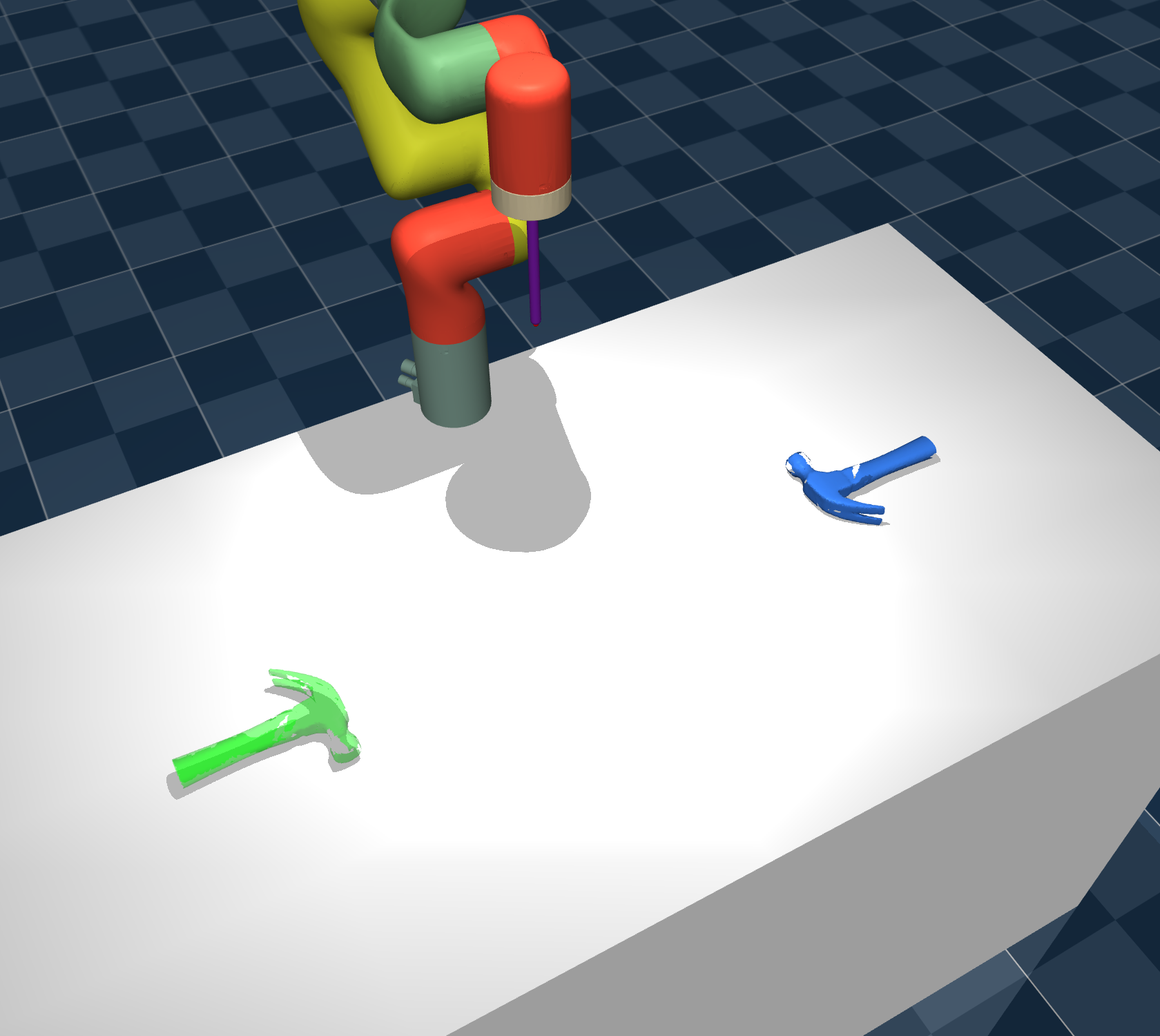}
        \subcaption{}\label{fig:tasks_sim_a}
    \end{minipage}\hfill
    \begin{minipage}[t]{0.16\linewidth}
        \centering
        \includegraphics[width=\linewidth]{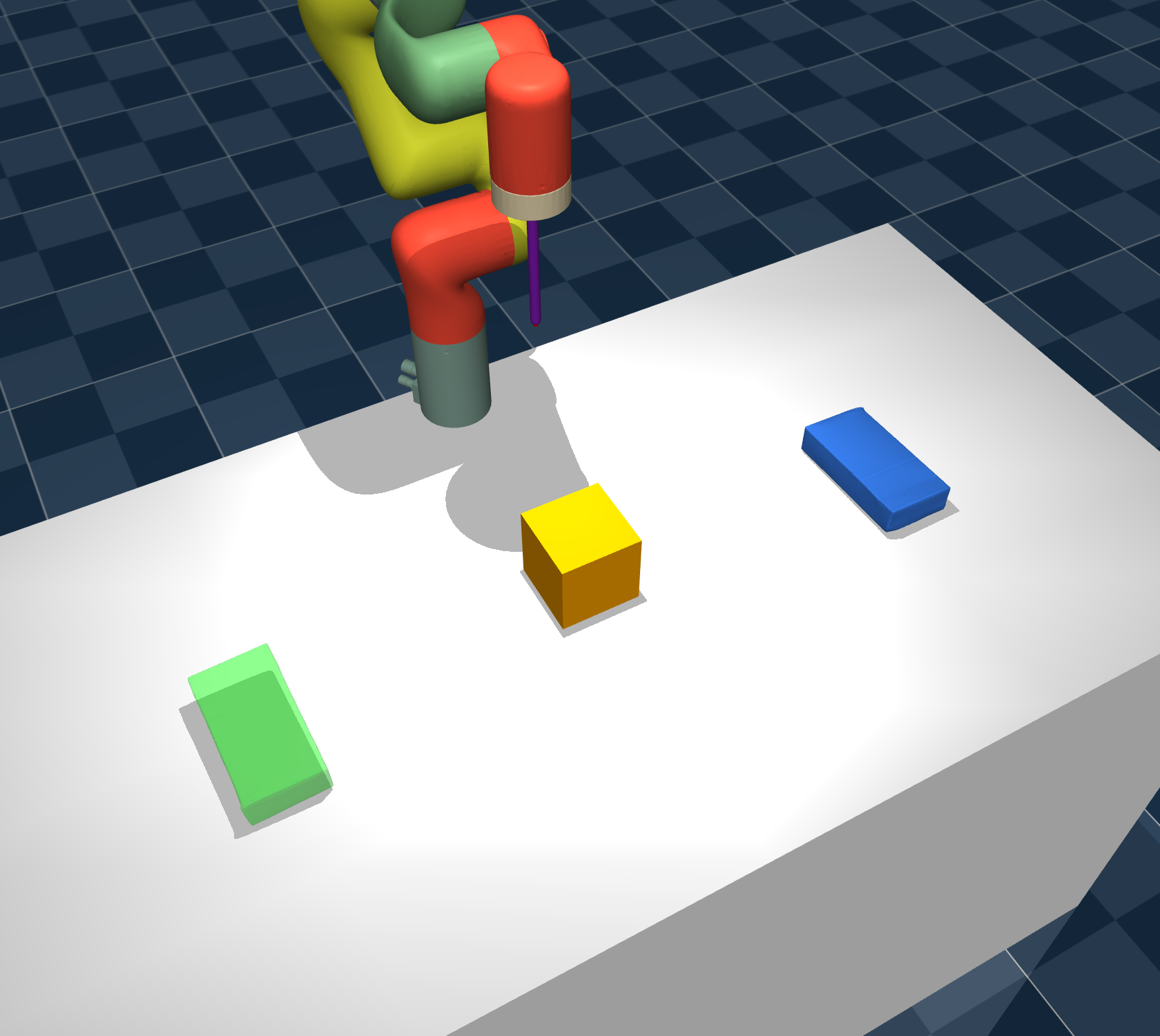}
        \subcaption{}\label{fig:tasks_sim_b}
    \end{minipage}\hfill
    \begin{minipage}[t]{0.16\linewidth}
        \centering
        \includegraphics[width=\linewidth]{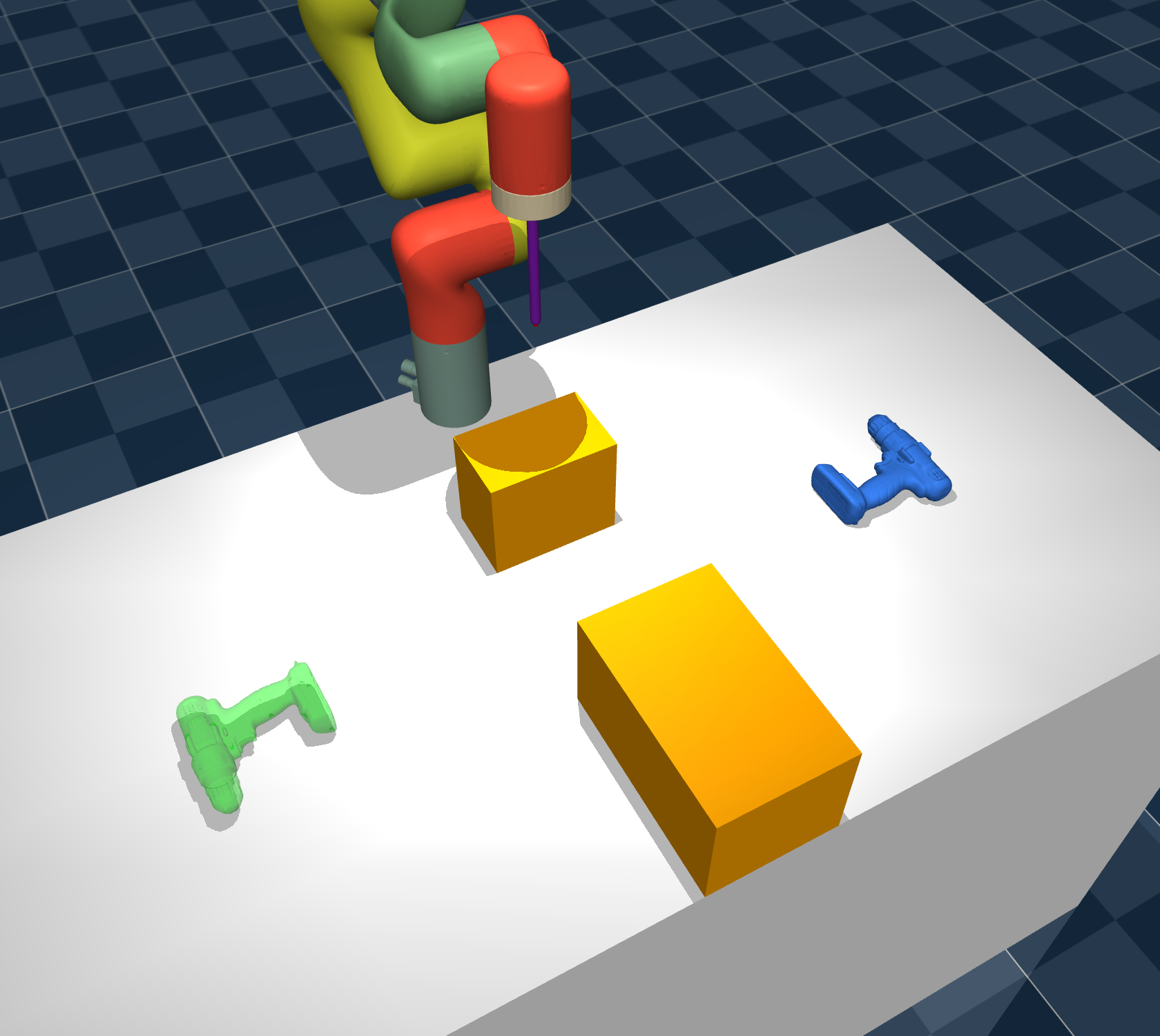}
        \subcaption{}\label{fig:tasks_sim_c}
    \end{minipage}\hfill
    \begin{minipage}[t]{0.16\linewidth}
        \centering
        \includegraphics[width=\linewidth]{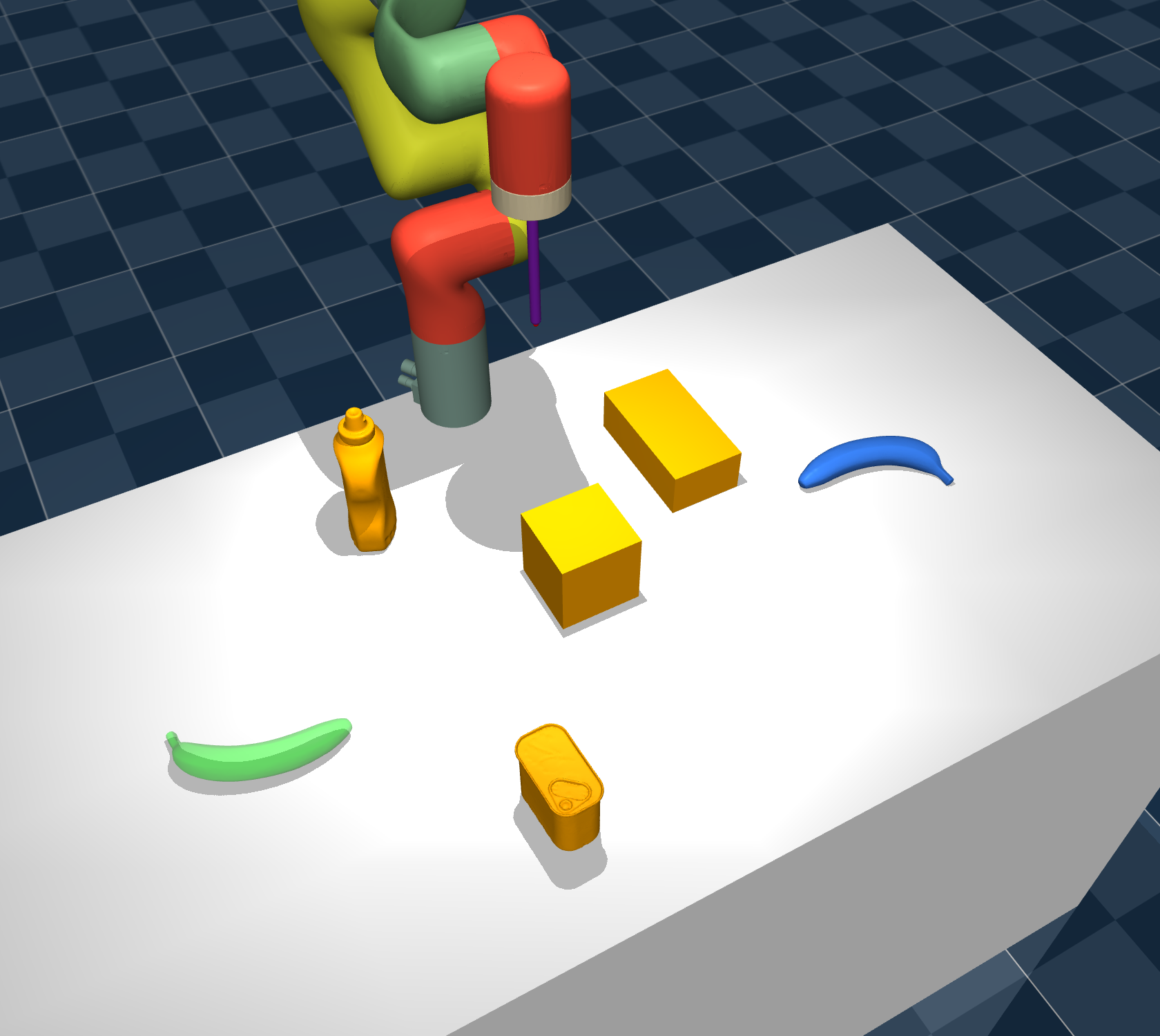}
        \subcaption{}\label{fig:tasks_sim_d}
    \end{minipage}\hfill
    \begin{minipage}[t]{0.16\linewidth}
        \centering
        \includegraphics[width=\linewidth]{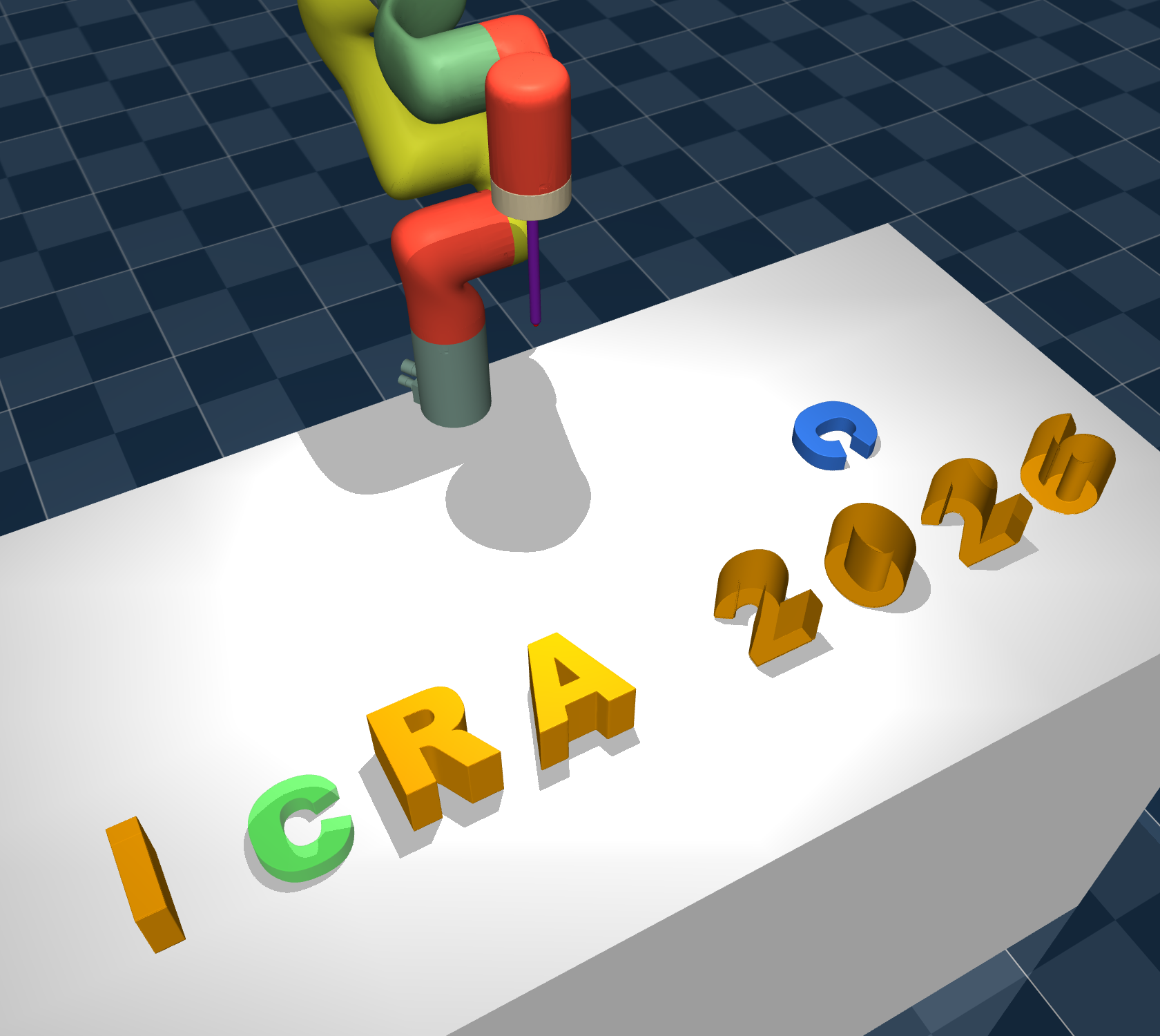}
        \subcaption{}\label{fig:tasks_sim_e}
    \end{minipage}\hfill
    \begin{minipage}[t]{0.16\linewidth}
        \centering
        \includegraphics[width=\linewidth]{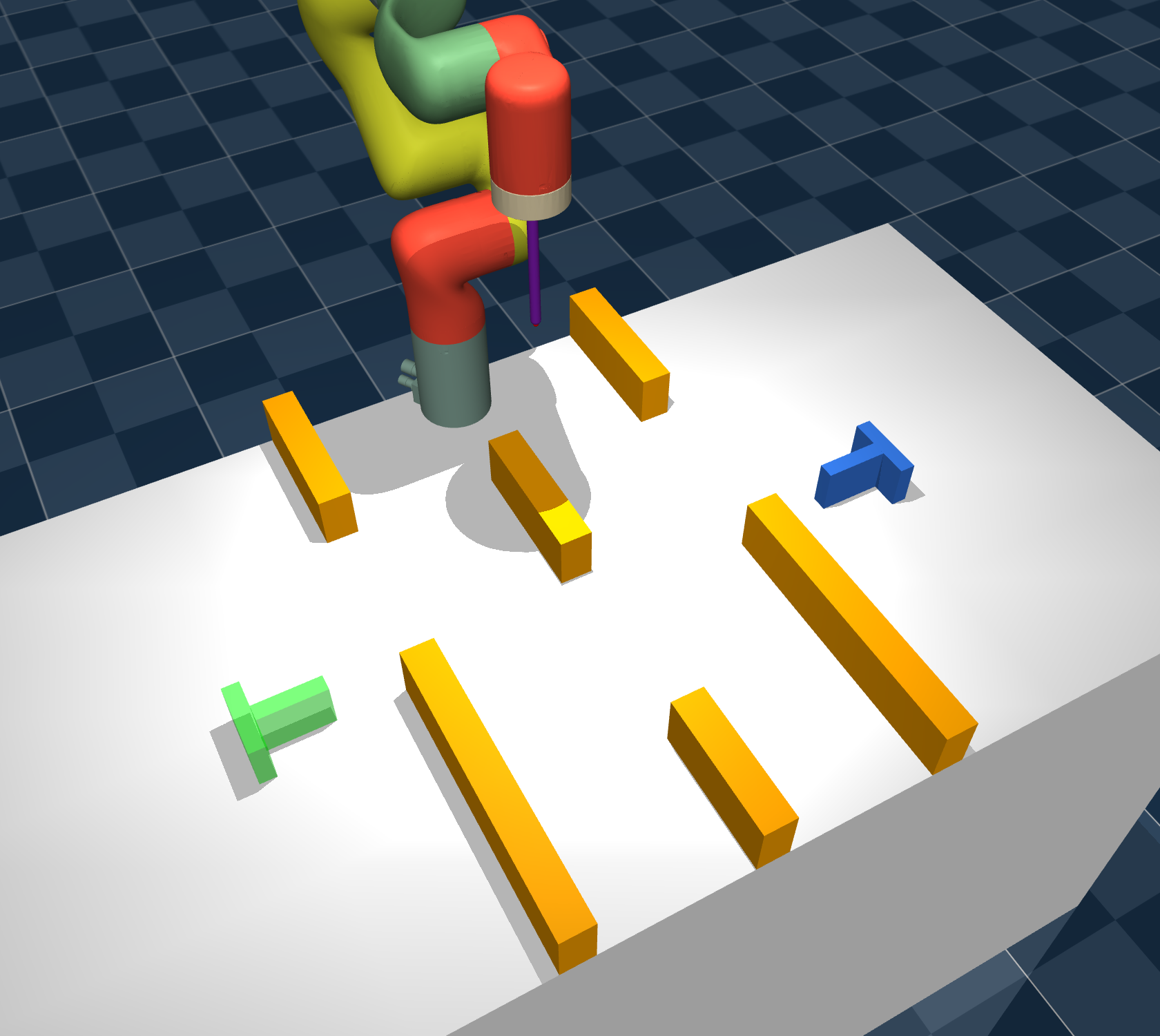}
        \subcaption{}\label{fig:tasks_sim_f}
    \end{minipage}
    \caption{\textbf{Simulation tasks.} (a) \emph{Open table}: a $180^\circ$ reorientation with no obstacles; (b) \emph{Single obstacle}: one cube on the direct path; (c) \emph{Shelf gap}: a gap as wide as the object is long; (d) \emph{Clutter}: a cube and three YCB objects; (e) \emph{Sign}: pushing the letter \emph{C} into place to complete ``ICRA''; (f) \emph{Slalom}: three gates whose openings alternate from one side of the table to the other. The object's initial pose is marked in blue, the goal as a translucent green outline, and obstacles in yellow.}
    \label{fig:tasks_sim}
\end{figure*}

\begin{table}[t]
\centering
\caption{\textbf{Parameter ablation.} CLOI is run on the T-shaped block with object pose as consensus variable, varying one parameter at a time about the default configuration ($\dag$). Every entry averages the six scenarios of Fig.~\ref{fig:tasks_sim} at 25 start--goal pairs each. Metrics are defined in Section~\ref{subsec:experiment_setup}.}
\label{tab:parameters}
\setlength{\tabcolsep}{3pt}
\footnotesize
\begin{tabular}{llccccccc}
\toprule
\cmidrule(lr){4-5} \cmidrule(lr){6-7}
Param & Value & SR $\uparrow$ & $\epsilon_d$ & $\epsilon_\theta$ & $\epsilon_d^s$ & $\epsilon_\theta^s$ & $T$ $\downarrow$ & $\bar{f}$ $\uparrow$ \\
\midrule
\multirow{3}{*}{$z_t$}
  & Pose$^\dag$ ($\lambda{=}0.1$)  & \textbf{0.96} & \textbf{0.48} & 1.12 & \textbf{0.48} & 1.12 & 21.5 & 1.2 \\
  & Wrench ($K{=}512$)         & 0.93 & 0.52 & \textbf{1.05} & 0.52 & \textbf{1.05} & 23.7 & \textbf{1.4} \\
  & Contact ($\lambda{=}0.25$) & 0.95 & 0.53 & 1.33 & 0.54 & 1.33 & \textbf{19.7} & \textbf{1.4} \\
\midrule
\multirow{4}{*}{$H^c$}
  & $16$                   & 0.87 & 0.48 & 1.20 & \textbf{0.46} & \textbf{1.10} & 26.4 & \textbf{2.2} \\
  & $24$                   & \textbf{0.94} & \textbf{0.47} & \textbf{1.11} & 0.48 & 1.13 & 21.8 & 1.5 \\
  & $32^{\dag}$            & \textbf{0.94} & 0.49 & 1.17 & 0.49 & 1.15 & \textbf{21.3} & 1.3 \\
  & $48$                   & 0.92 & 0.49 & 1.18 & 0.49 & 1.22 & 22.7 & 0.8 \\
\midrule
\multirow{4}{*}{$K$}
  & $64$                   & 0.87 & \textbf{0.47} & \textbf{1.11} & \textbf{0.49} & 1.14 & 26.3 & \textbf{1.4} \\
  & $128$                  & 0.91 & 0.49 & 1.21 & 0.49 & 1.19 & 24.5 & 1.3 \\
  & $256^{\dag}$           & 0.94 & 0.49 & 1.17 & 0.49 & 1.15 & 21.3 & 1.3 \\
  & $512$                  & \textbf{0.96} & 0.49 & 1.11 & 0.49 & \textbf{1.12} & \textbf{20.9} & 1.1 \\
\midrule
\multirow{4}{*}{$N_{\mathrm{ADMM}}$}
  & $1$                    & 0.80 & \textbf{0.46} & \textbf{0.83} & \textbf{0.44} & \textbf{0.84} & 31.8 & \textbf{3.6} \\
  & $2$                    & 0.88 & 0.49 & 1.01 & 0.49 & 1.06 & 24.8 & 2.0 \\
  & $4^{\dag}$             & 0.94 & 0.49 & 1.17 & 0.49 & 1.15 & \textbf{21.3} & 1.3 \\
  & $8$                    & \textbf{0.95} & 0.49 & 1.20 & 0.49 & 1.18 & 21.6 & 0.6 \\
\midrule
\multirow{4}{*}{$\rho$}
  & $1.0$                  & \textbf{0.94} & 0.52 & 1.25 & 0.52 & 1.26 & \textbf{20.6} & 1.1 \\
  & $2.0^{\dag}$           & \textbf{0.94} & 0.49 & 1.17 & 0.49 & 1.15 & 21.3 & \textbf{1.3} \\
  & $4.0$                  & 0.89 & \textbf{0.45} & 0.98 & 0.45 & 1.01 & 26.6 & 1.2 \\
  & $8.0$                  & 0.84 & 0.46 & \textbf{0.91} & \textbf{0.45} & \textbf{0.95} & 31.1 & 1.2 \\
\midrule
\multirow{4}{*}{$\nu$}
  & $0.2$                  & 0.74 & 0.52 & \textbf{0.70} & 0.51 & \textbf{0.76} & 33.4 & 1.2 \\
  & $0.4$                  & 0.82 & 0.50 & 0.73 & 0.49 & 0.78 & 31.0 & 1.2 \\
  & $0.6$                  & 0.89 & \textbf{0.49} & 0.84 & \textbf{0.48} & 0.87 & 26.9 & 1.2 \\
  & $0.8^{\dag}$           & \textbf{0.94} & 0.49 & 1.17 & 0.49 & 1.15 & \textbf{21.3} & \textbf{1.3} \\
\midrule
\multirow{4}{*}{$\lambda$}
  & $0.10$                 & \textbf{0.96} & 0.48 & 1.12 & 0.48 & 1.12 & 21.5 & 1.2 \\
  & $0.25$                 & 0.94 & 0.49 & 1.15 & 0.48 & 1.16 & 21.6 & 1.2 \\
  & $0.50^{\dag}$          & 0.94 & 0.49 & 1.17 & 0.49 & 1.15 & \textbf{21.3} & \textbf{1.3} \\
  & $1.00$                 & 0.87 & \textbf{0.47} & \textbf{1.07} & \textbf{0.48} & \textbf{1.11} & 24.4 & 1.2 \\
\bottomrule
\end{tabular}
\end{table}

\begin{table*}[t]
\centering
\caption{\textbf{Baseline Comparison.} Every method at its own best configuration, on six scenarios with five pushed objects (T-block, hammer, sugar box, power drill, banana) across 25 different start--goal positions, total of $125$ trials per scenario.}
\label{tab:ablation}
\setlength{\tabcolsep}{5pt}
\resizebox{\linewidth}{!}{\begin{tabular}{lcccccc c ccccccc}
\toprule
& \multicolumn{6}{c}{Success rate $\uparrow$ per scenario} & & \multicolumn{7}{c}{Mean over scenarios} \\
\cmidrule(lr){2-7} \cmidrule(lr){9-15}
Method & Open table & Single obst. & Shelf gap & Clutter & Sign & Slalom & & SR $\uparrow$ & $\epsilon_d$ $\downarrow$ & $\epsilon_\theta$ $\downarrow$ & $\epsilon_d^s$ $\downarrow$ & $\epsilon_\theta^s$ $\downarrow$ & $T$ $\downarrow$ & $\bar{f}$ $\uparrow$ \\
\midrule
MPPI~\cite{pezzato2025samplingbasedmodelpredictivecontrol} ($\lambda{=}0.1$) & 0.41 & 0.40 & 0.23 & 0.30 & 0.11 & 0.03 &  & 0.25 & 0.512 & 1.457 & 0.447 & 1.406 & 43.5 & 9.5 \\
MPPI (itr=8, $\lambda{=}0.1$) & 0.50 & 0.47 & 0.24 & 0.38 & 0.08 & 0.05 &  & 0.29 & 0.524 & 1.355 & 0.501 & 1.544 & 41.0 & 1.3 \\
C3+~\cite{bui2025pushanythingsinglemultiobject} & 0.66 & 0.24 & 0.18 & 0.14 & 0.07 & 0.00 &  & 0.21 & \textbf{0.461} & 1.463 & \textbf{0.348} & \textbf{0.971} & 267.2 & \textbf{20.2} \\
\midrule
CLOI (object pose) ($\lambda{=}0.1$) & \textbf{0.97} & \textbf{0.90} & \textbf{0.58} & \textbf{0.61} & \textbf{0.61} & \textbf{0.20} &  & \textbf{0.64} & 0.470 & \textbf{1.233} & 0.450 & 1.296 & \textbf{33.3} & 1.1 \\
\bottomrule
\end{tabular}}
\end{table*}

\begin{figure*}[thpb]
    \centering

    \setlength{\tabcolsep}{0pt}    
    \renewcommand{\arraystretch}{1}

    \newcommand{\rowlab}[1]{\raisebox{2.5ex}{\rotatebox{90}{\scriptsize #1}}}
    \newcommand{\colgap}{\hspace{1.5pt}}

    \begin{tabular}{@{}c@{\colgap}c@{\colgap}c@{\colgap}c@{\colgap}c@{\colgap}c@{\colgap}c@{}}
        \multicolumn{7}{c}{\textbf{Successful Trials}} \\[1pt]
        
        \rowlab{CLOI} &
        \includegraphics[width=0.16\linewidth]{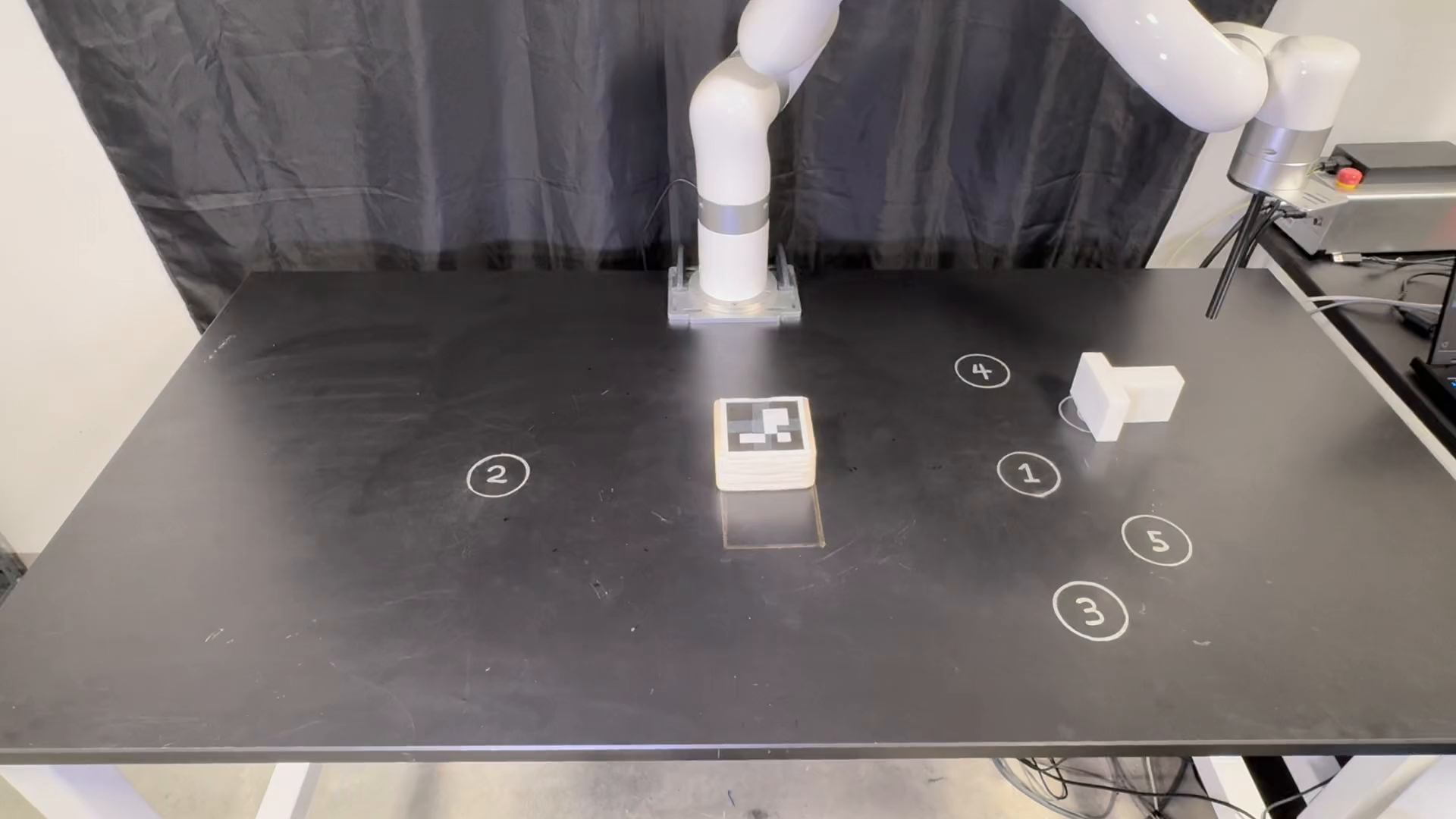} &
        \includegraphics[width=0.16\linewidth]{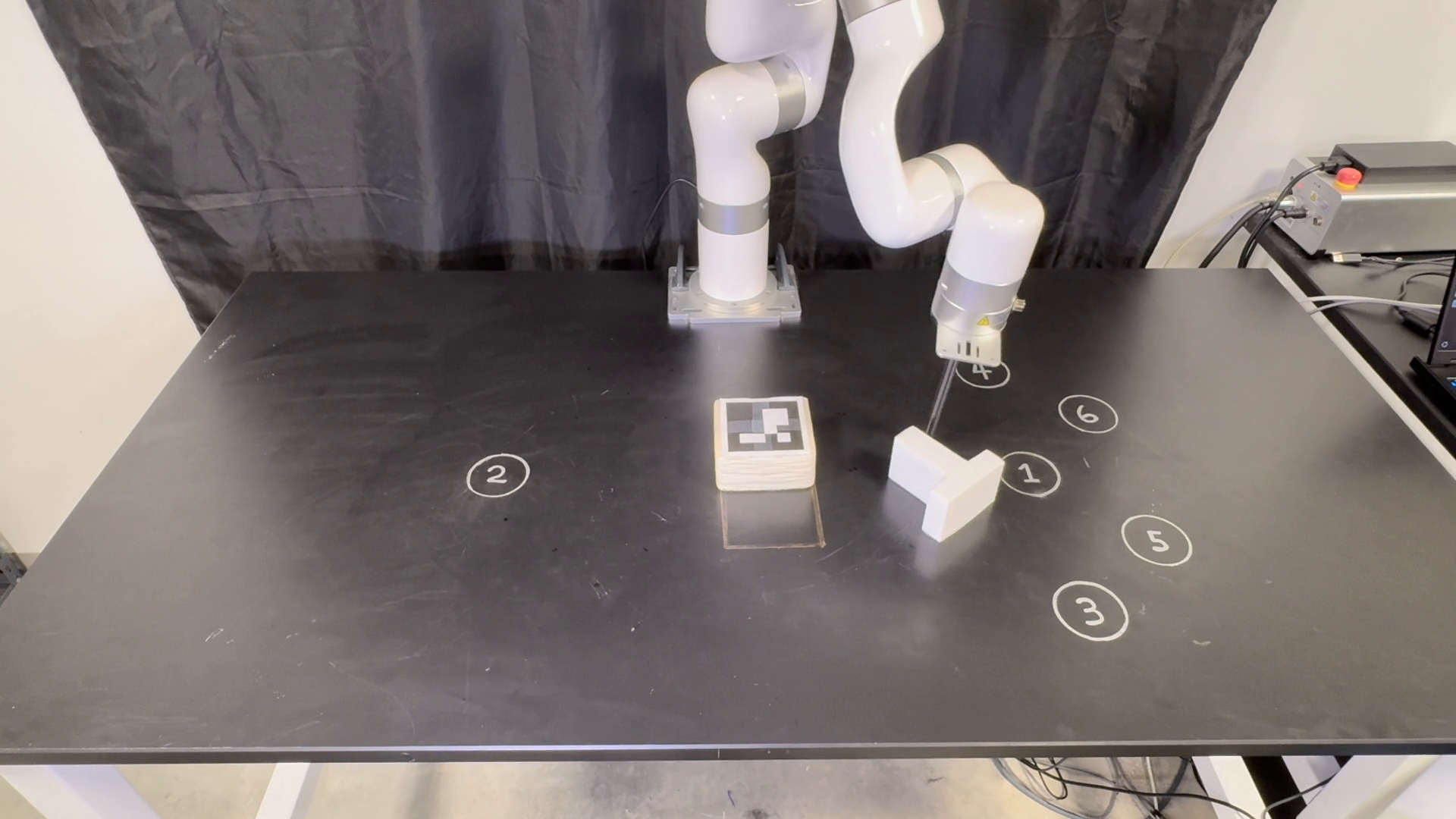} &
        \includegraphics[width=0.16\linewidth]{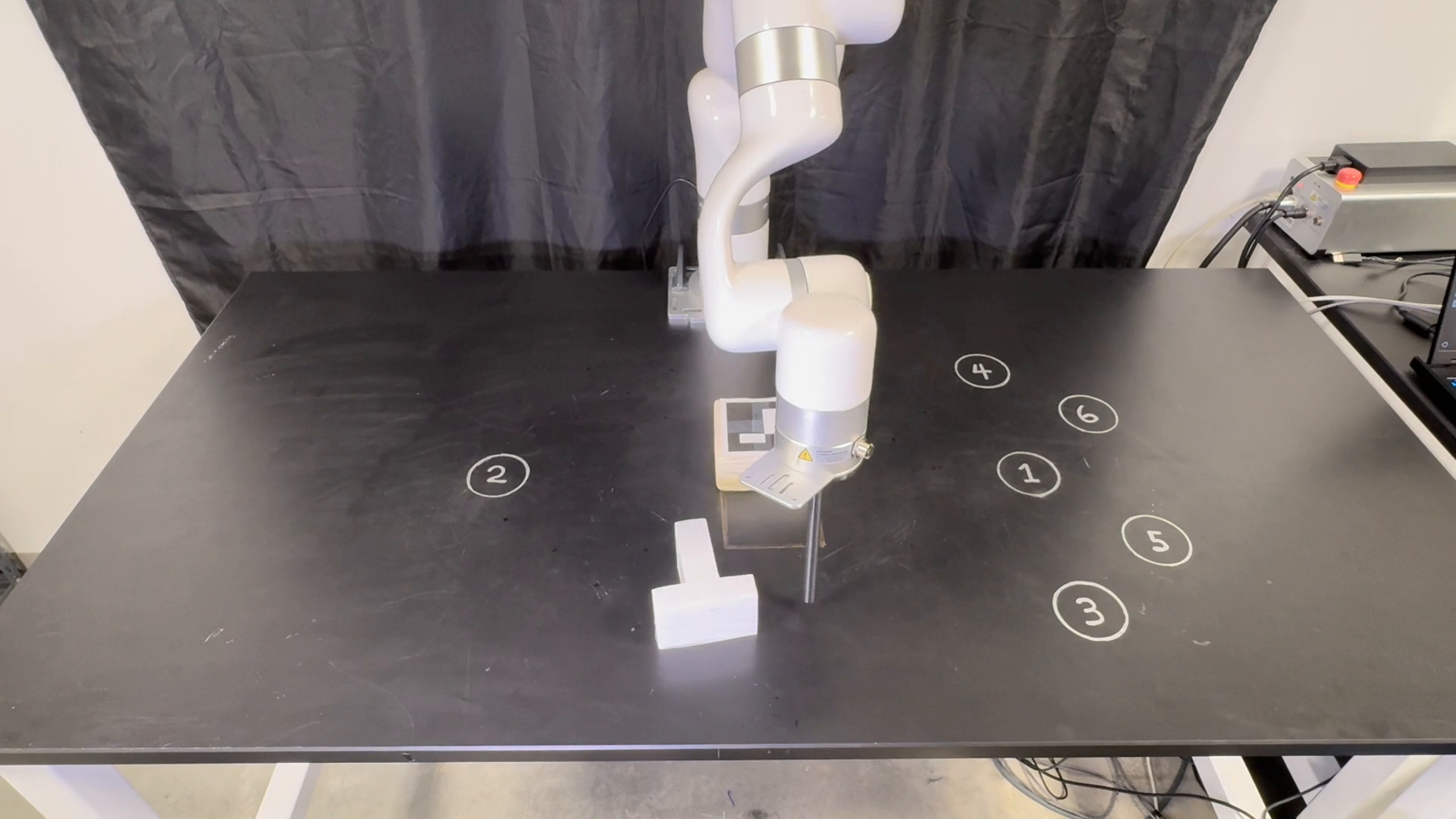} &
        \includegraphics[width=0.16\linewidth]{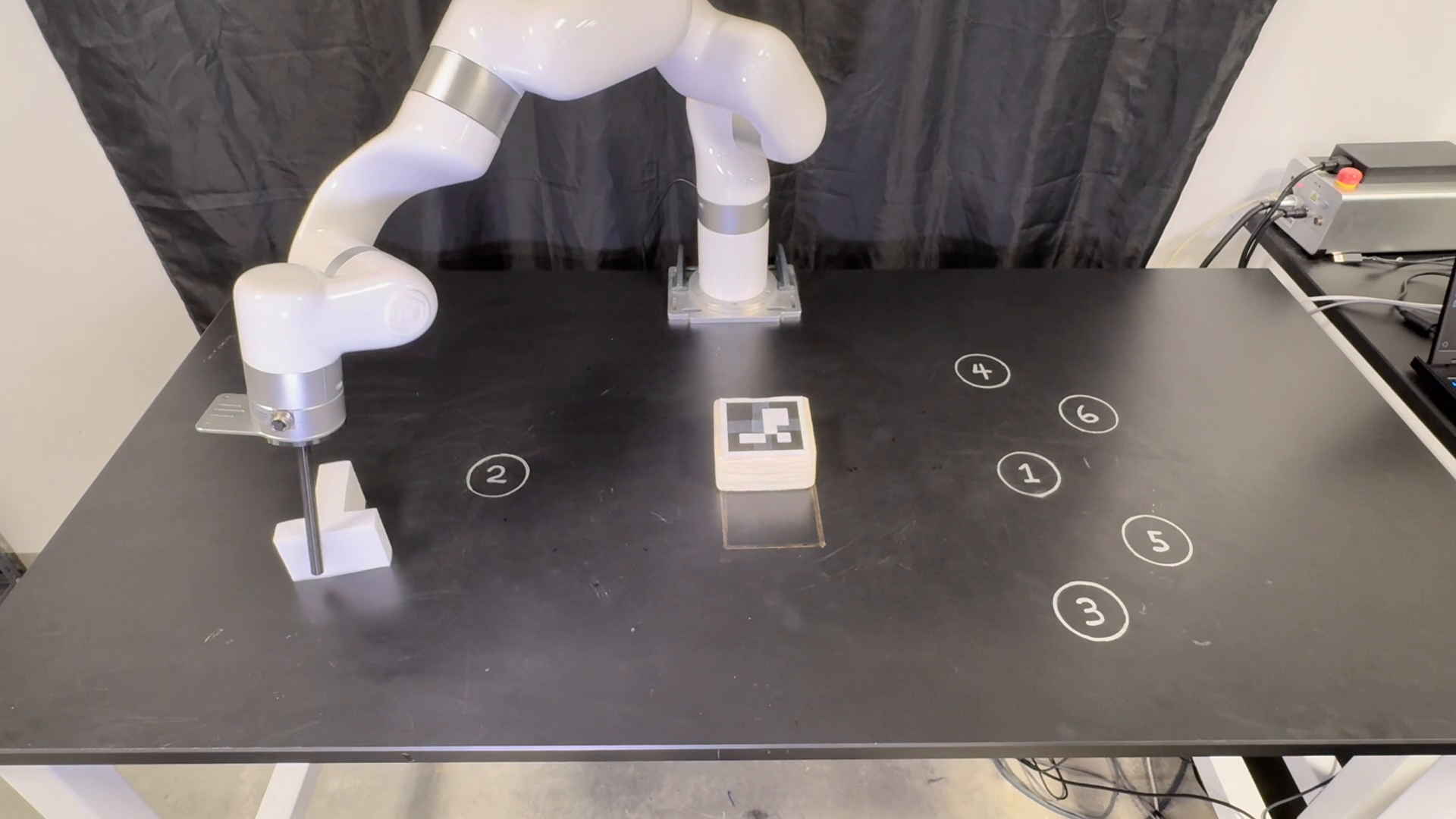} &
        \includegraphics[width=0.16\linewidth]{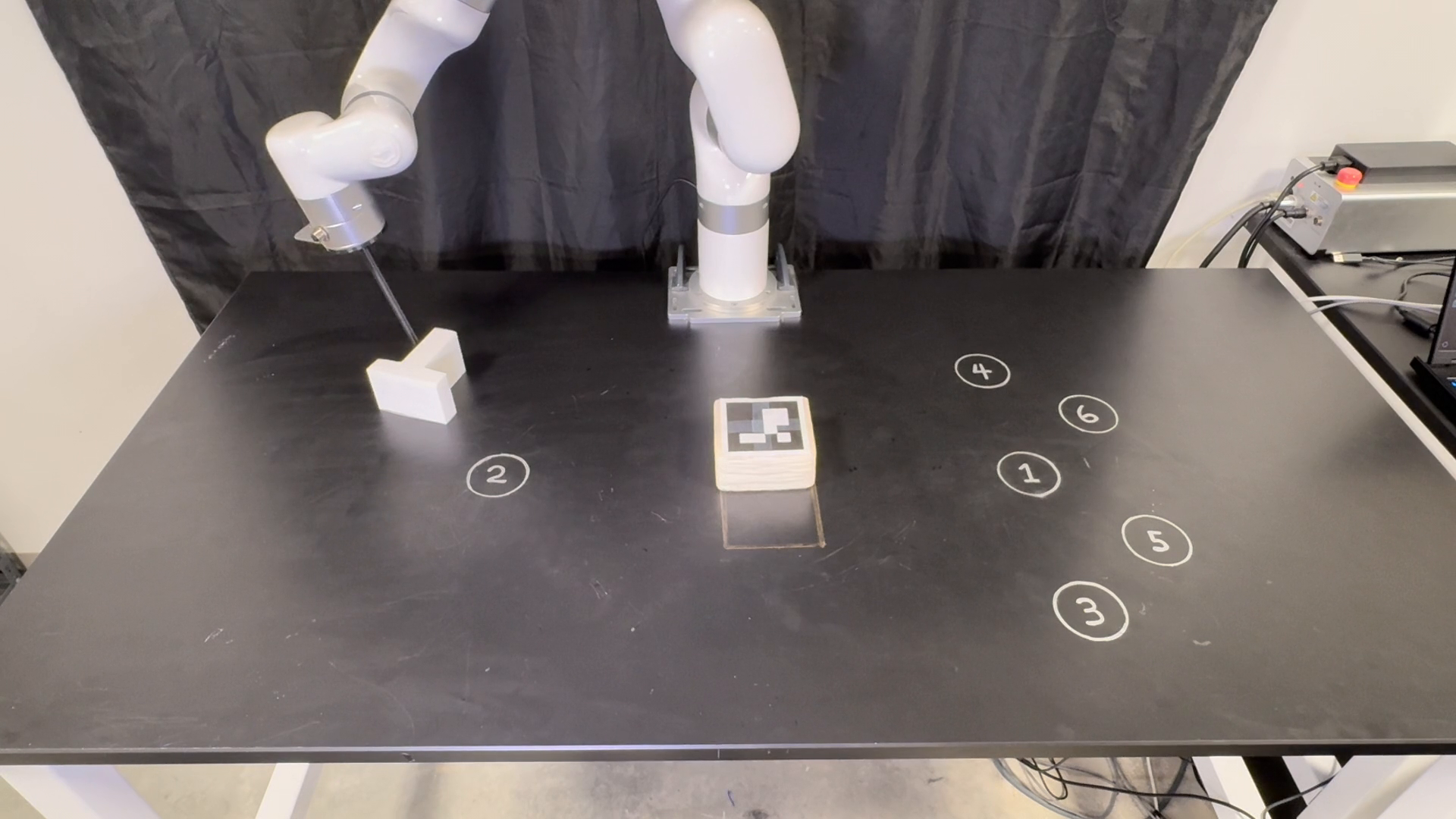} &
        \includegraphics[width=0.16\linewidth]{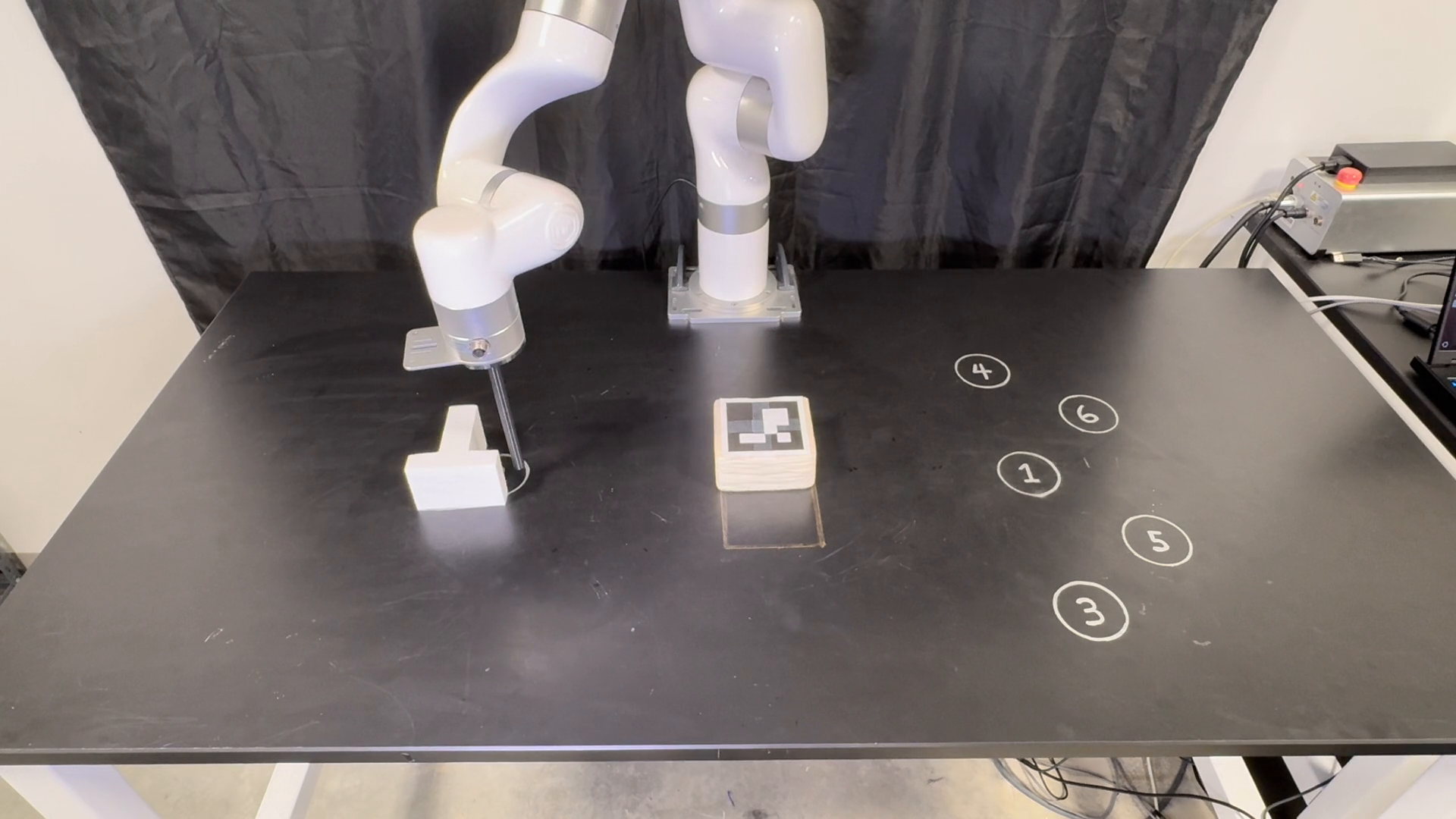} \\[-2pt]

        \rowlab{MPPI} &
        \includegraphics[width=0.16\linewidth]{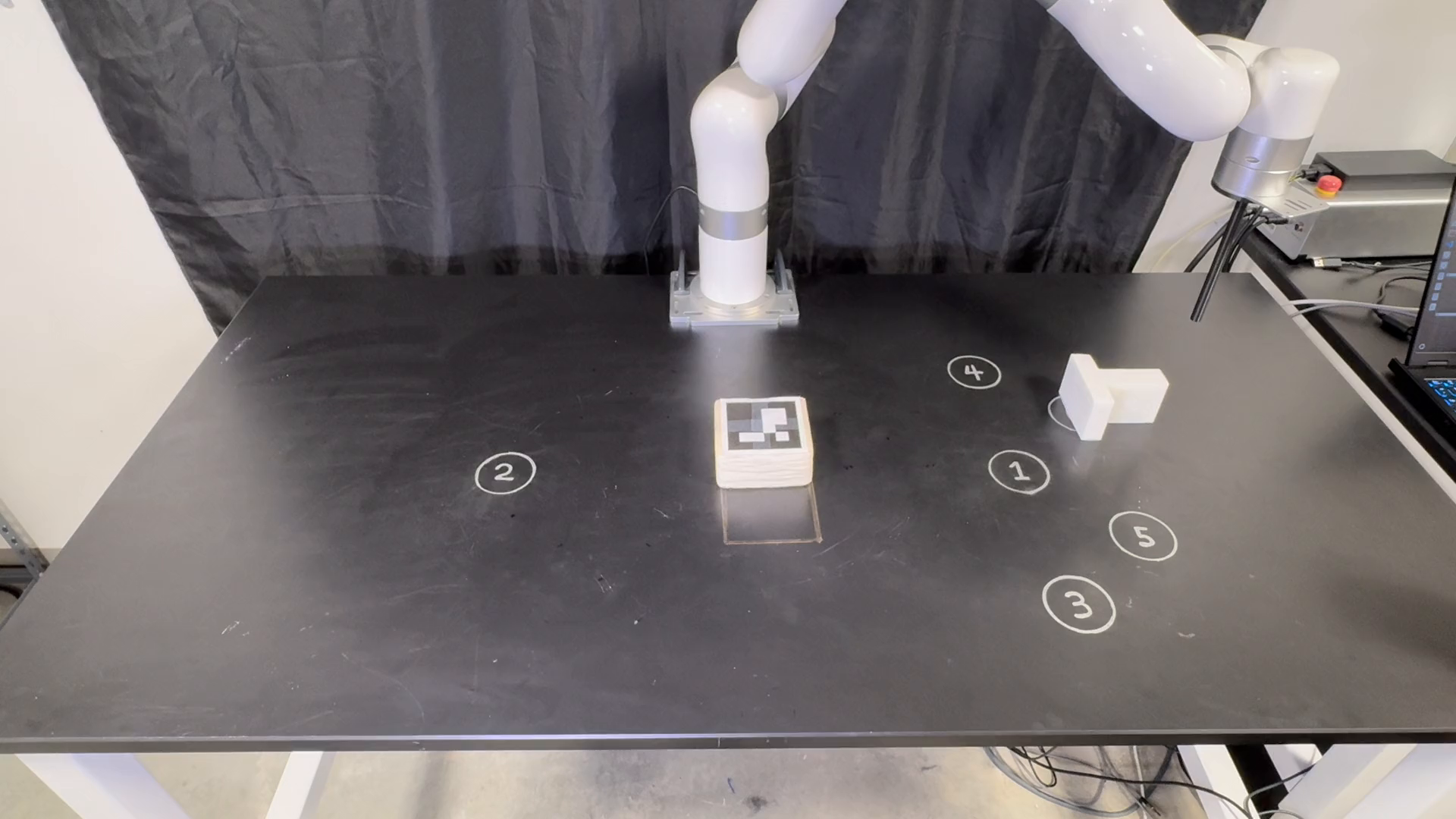} &
        \includegraphics[width=0.16\linewidth]{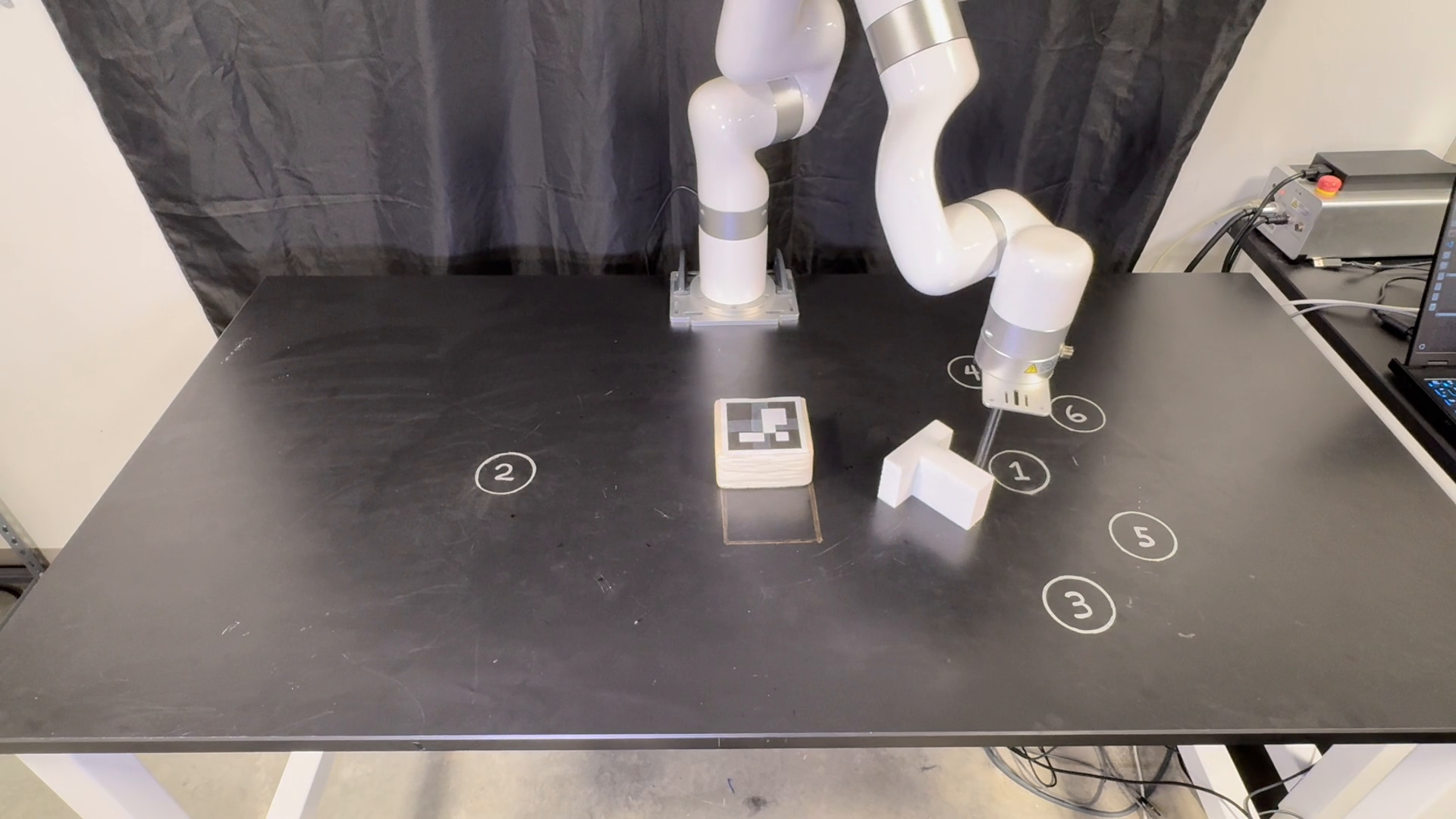} &
        \includegraphics[width=0.16\linewidth]{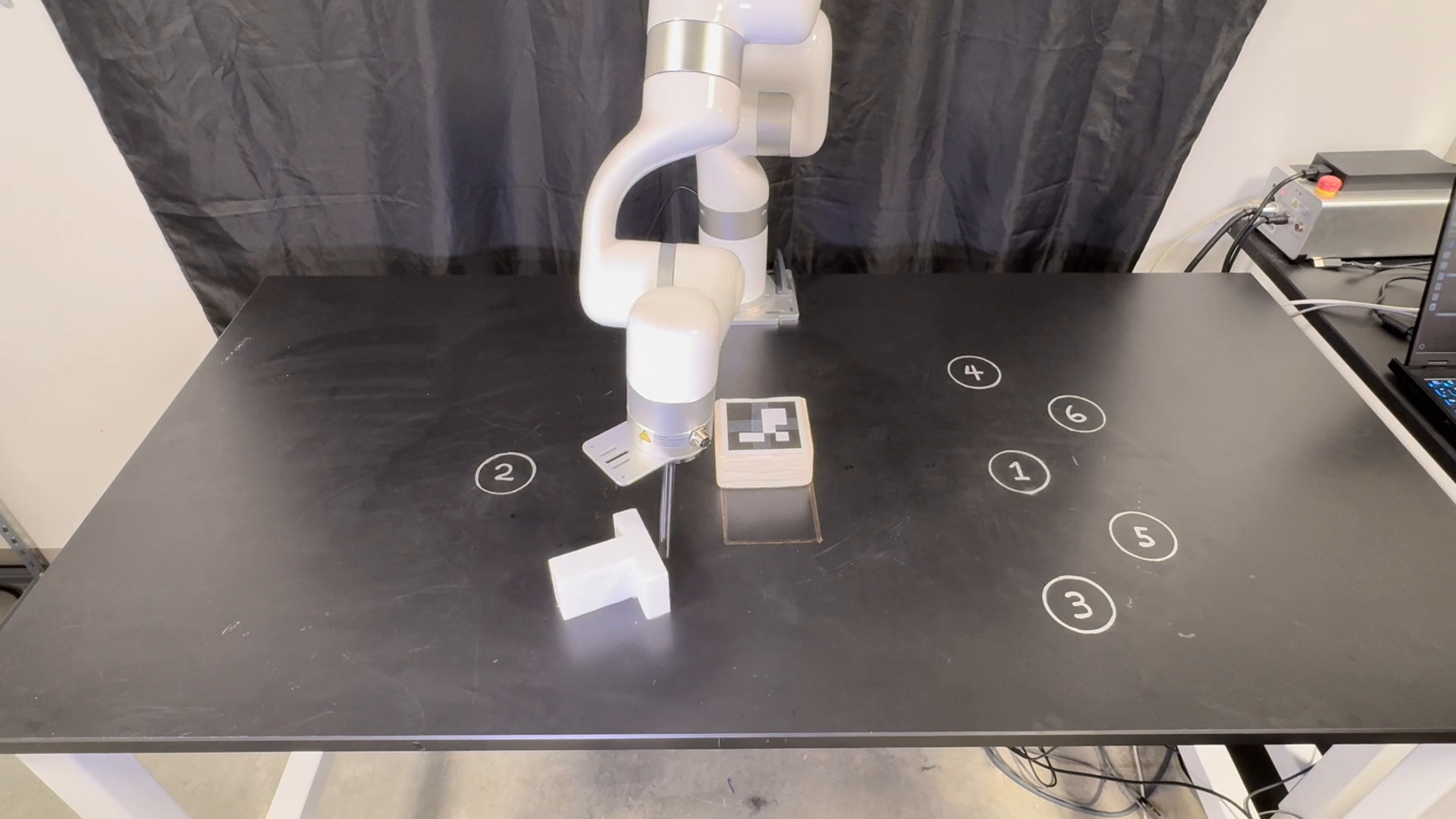} &
        \includegraphics[width=0.16\linewidth]{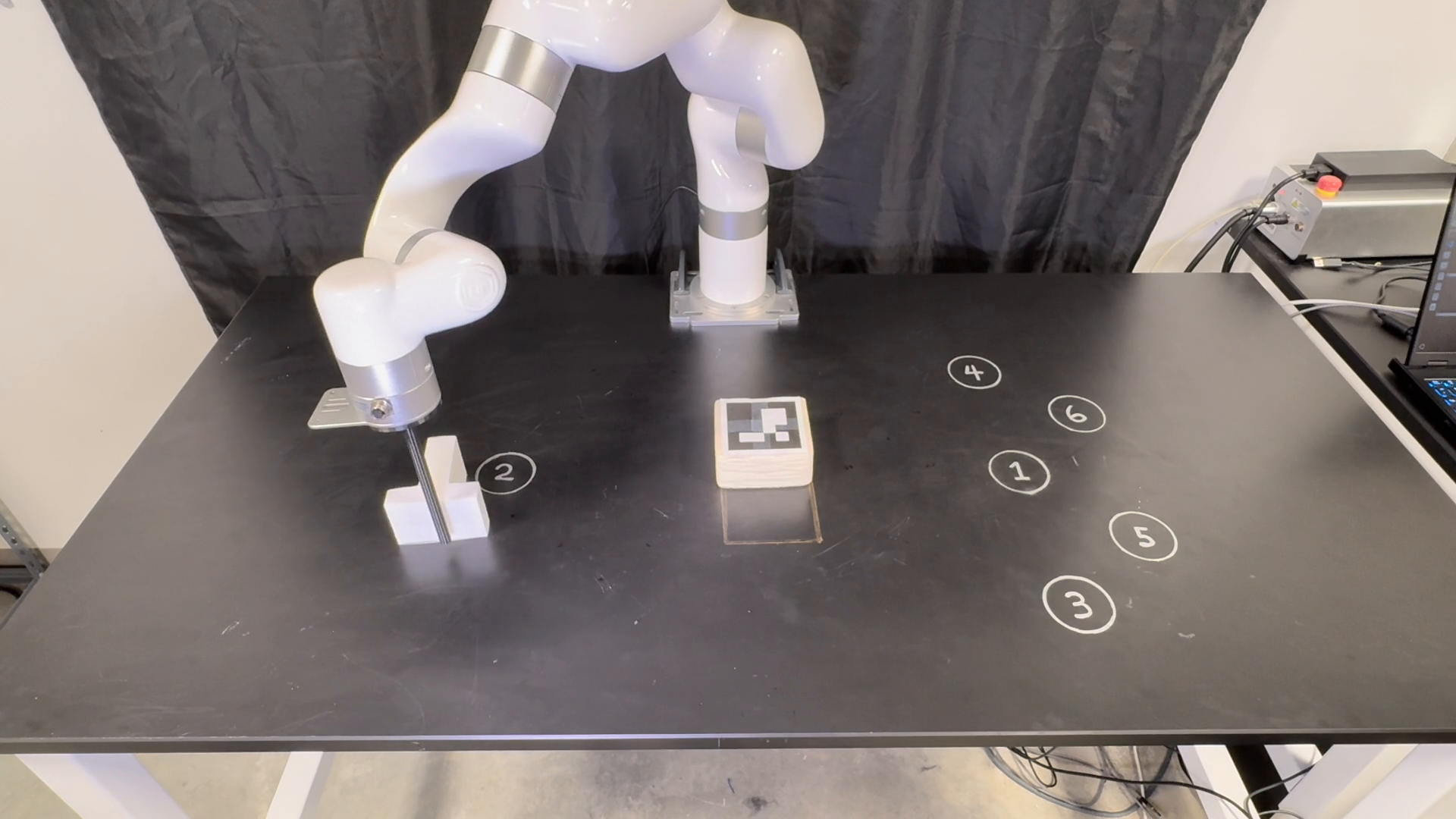} &
        \includegraphics[width=0.16\linewidth]{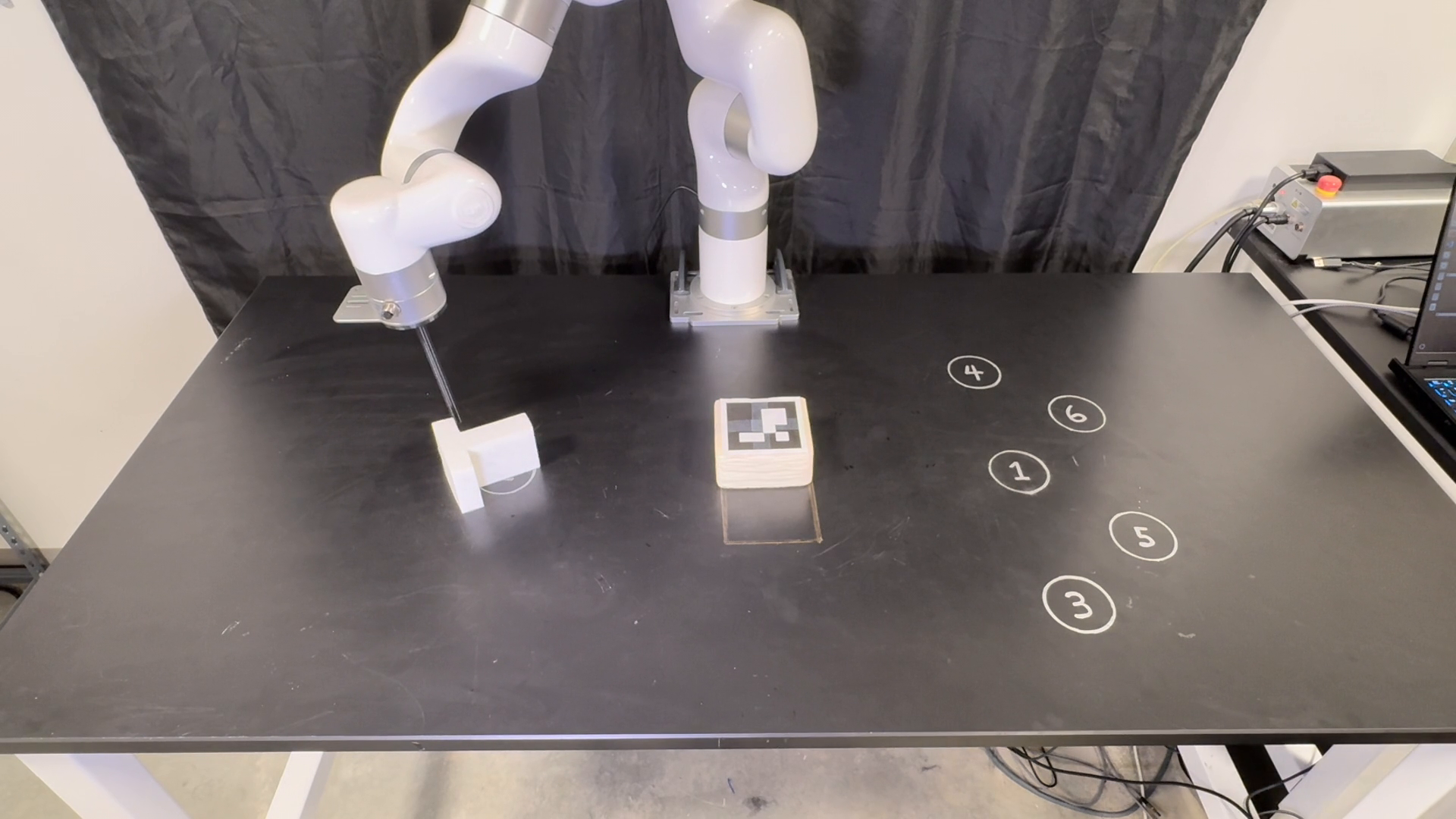} &
        \includegraphics[width=0.16\linewidth]{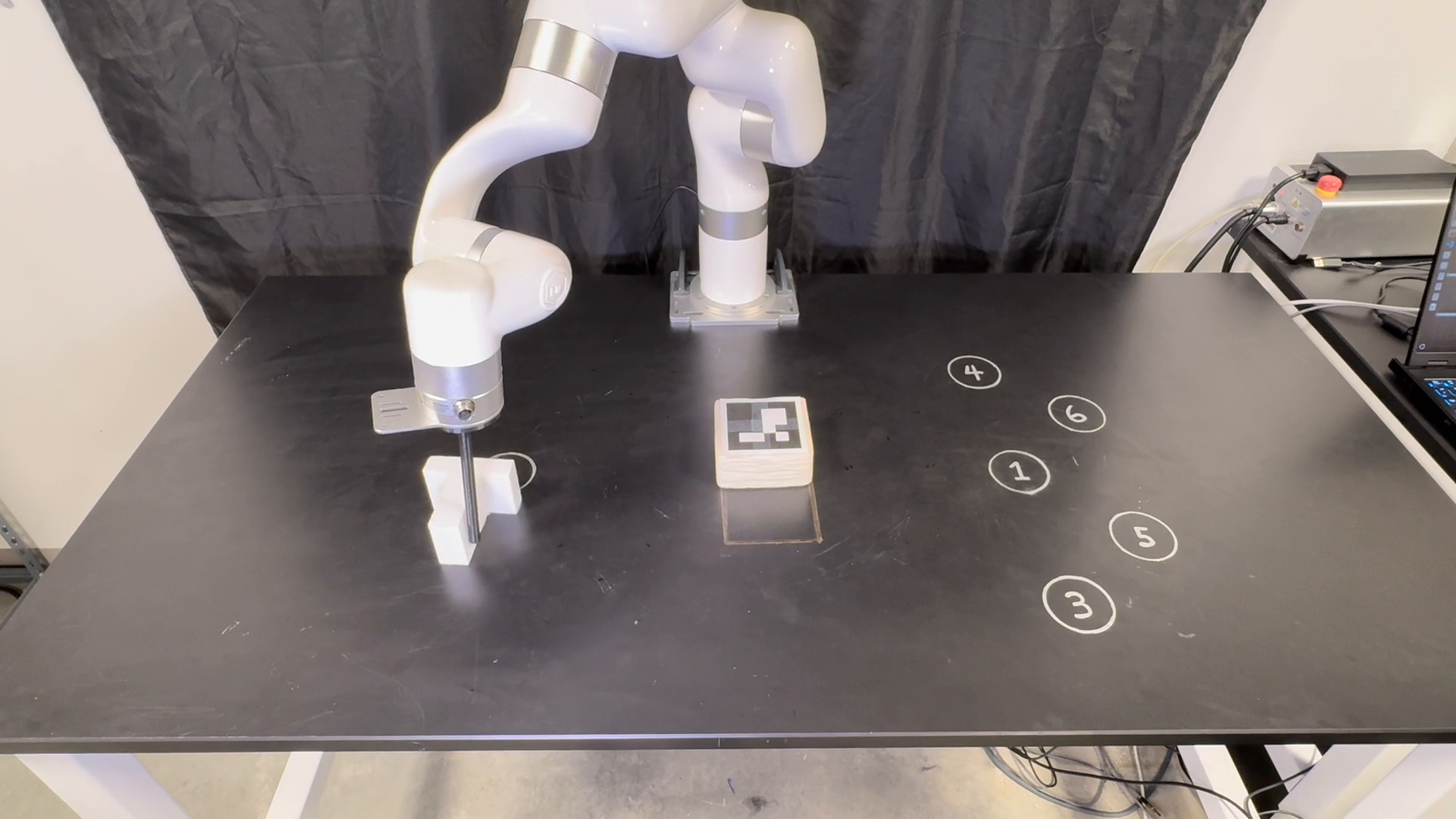} \\

        \multicolumn{7}{c}{\textbf{Failed Trials}} \\[2pt]

        \rowlab{CLOI} &
        \includegraphics[width=0.16\linewidth]{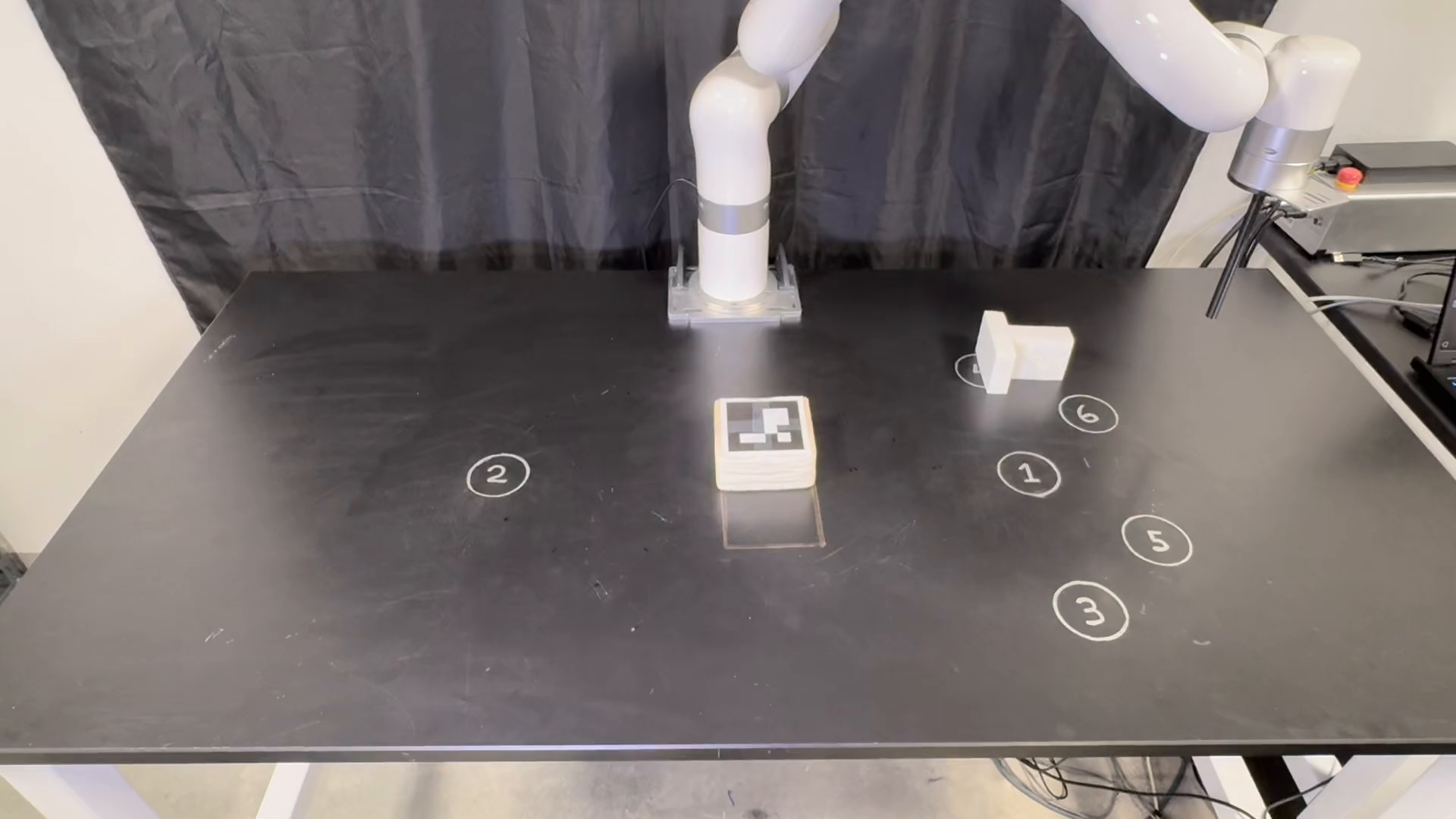} &
        \includegraphics[width=0.16\linewidth]{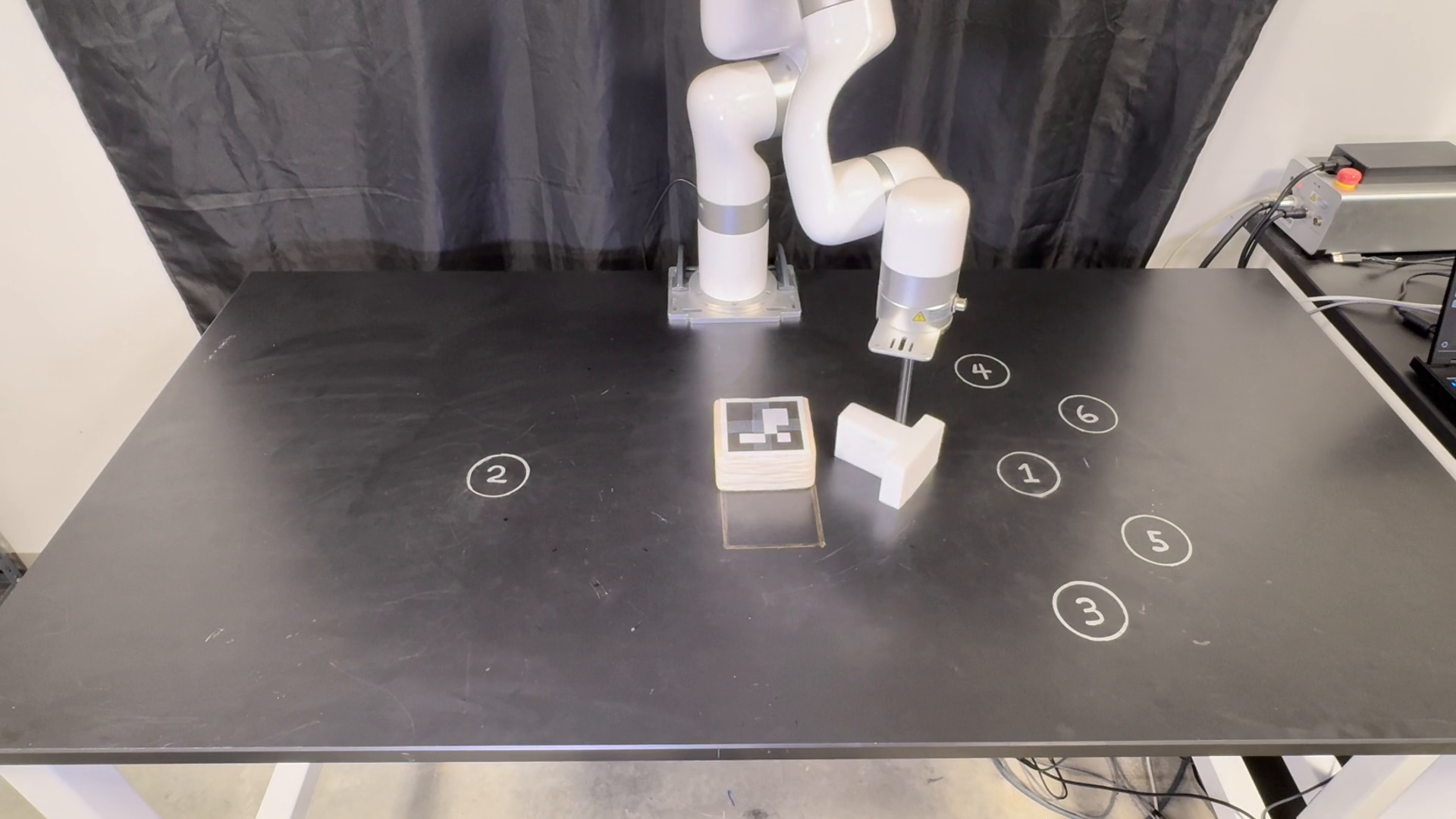} &
        \includegraphics[width=0.16\linewidth]{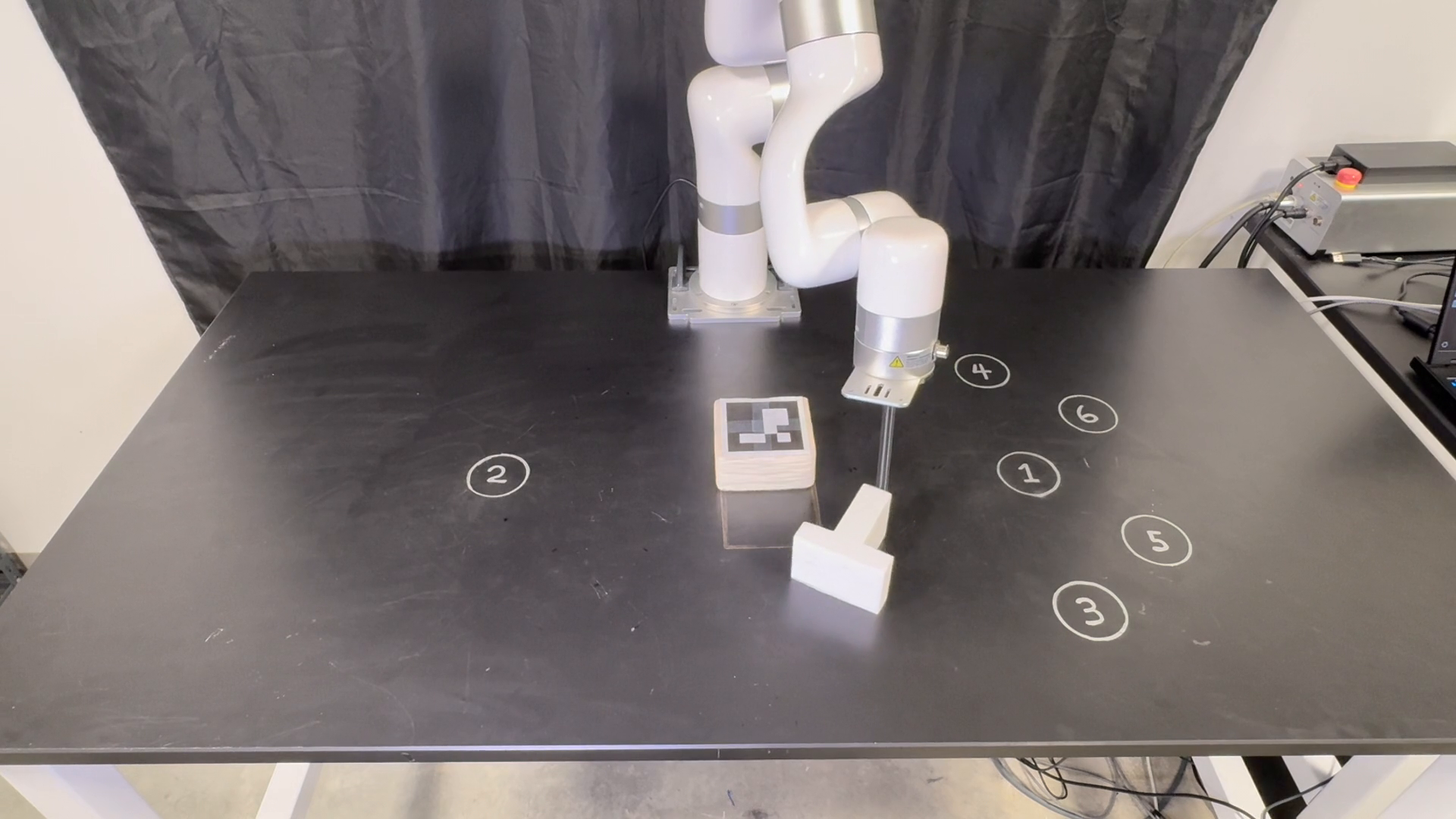} &
        \includegraphics[width=0.16\linewidth]{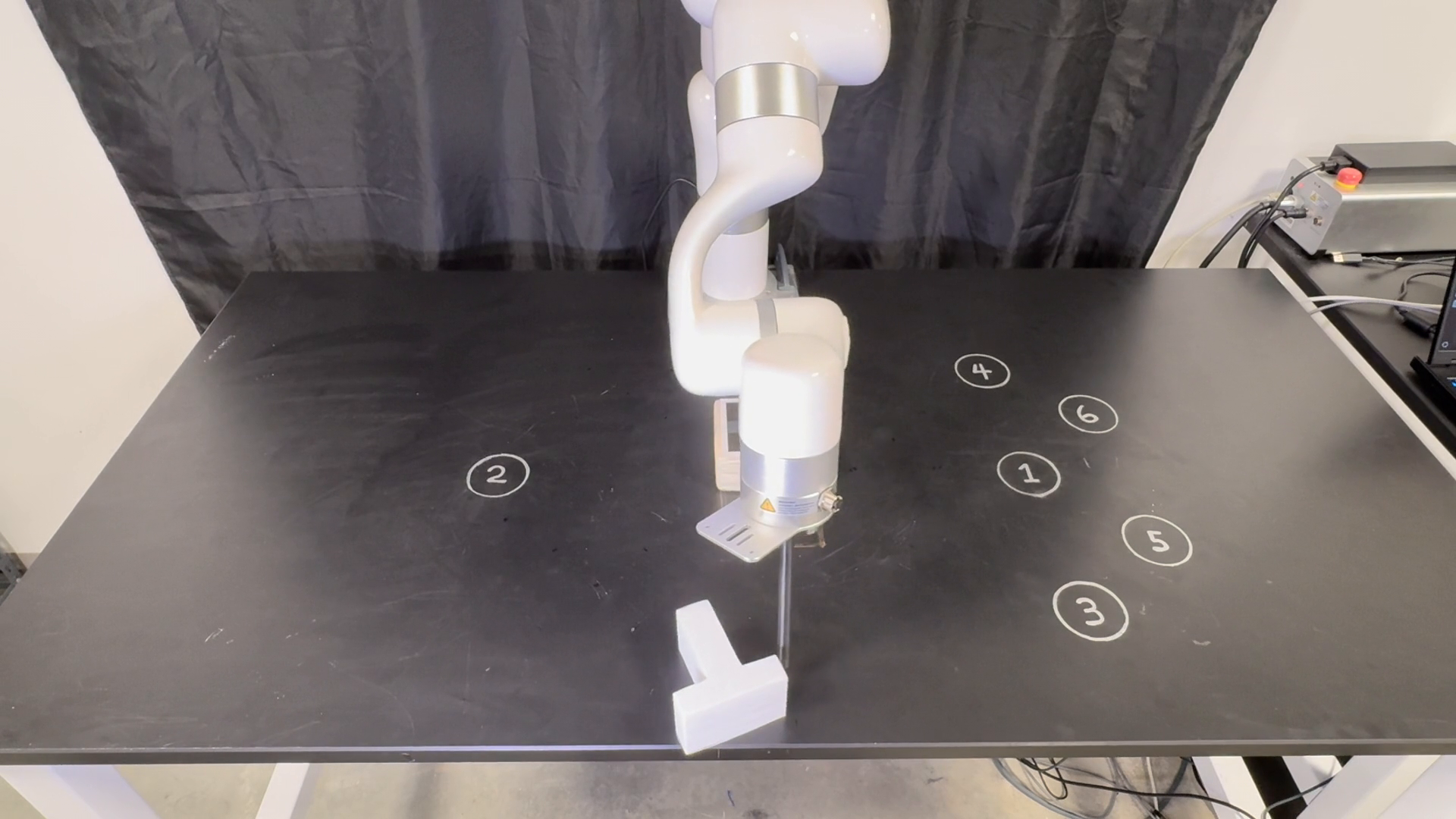} &
        \includegraphics[width=0.16\linewidth]{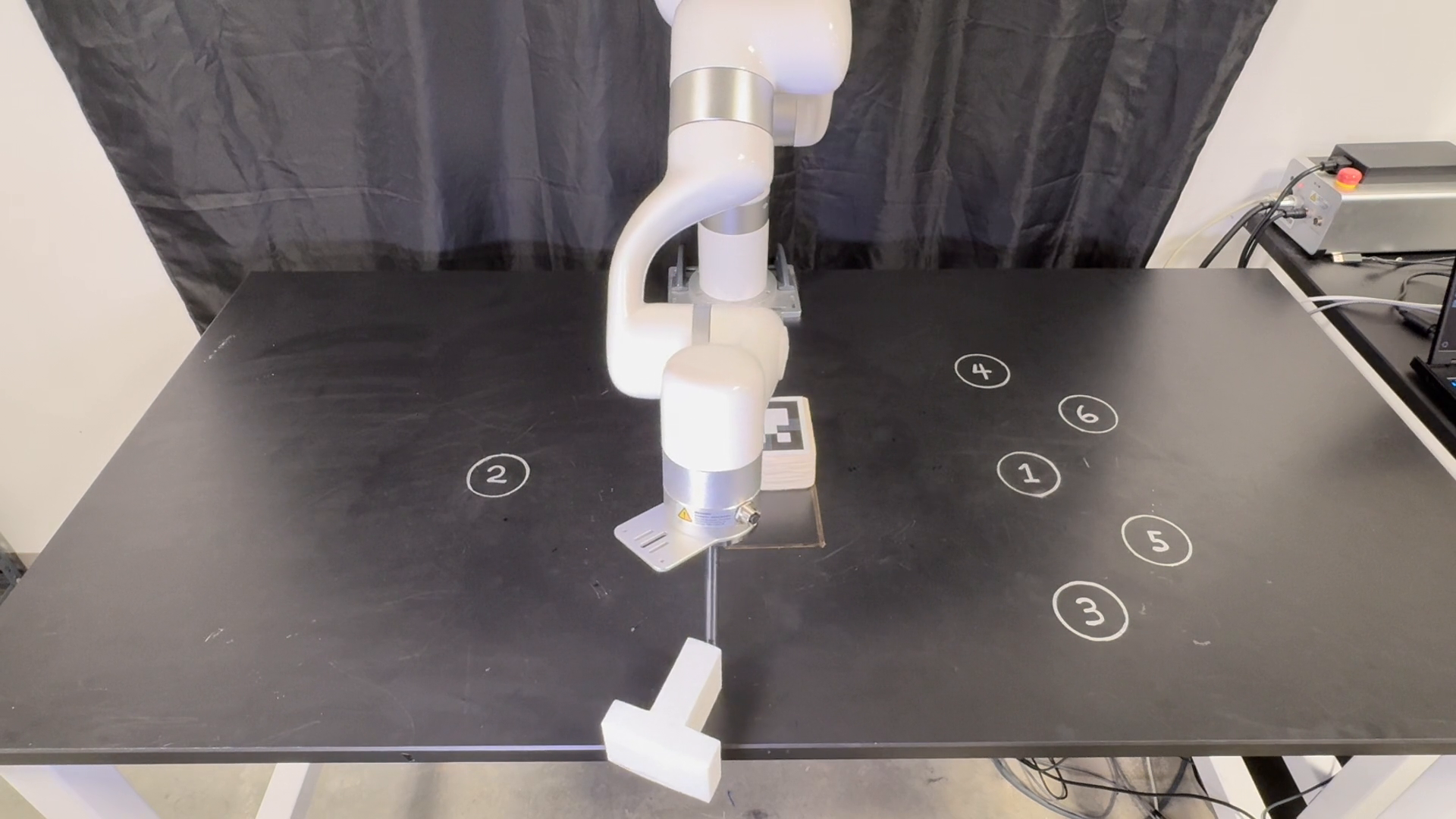} &
        \includegraphics[width=0.16\linewidth]{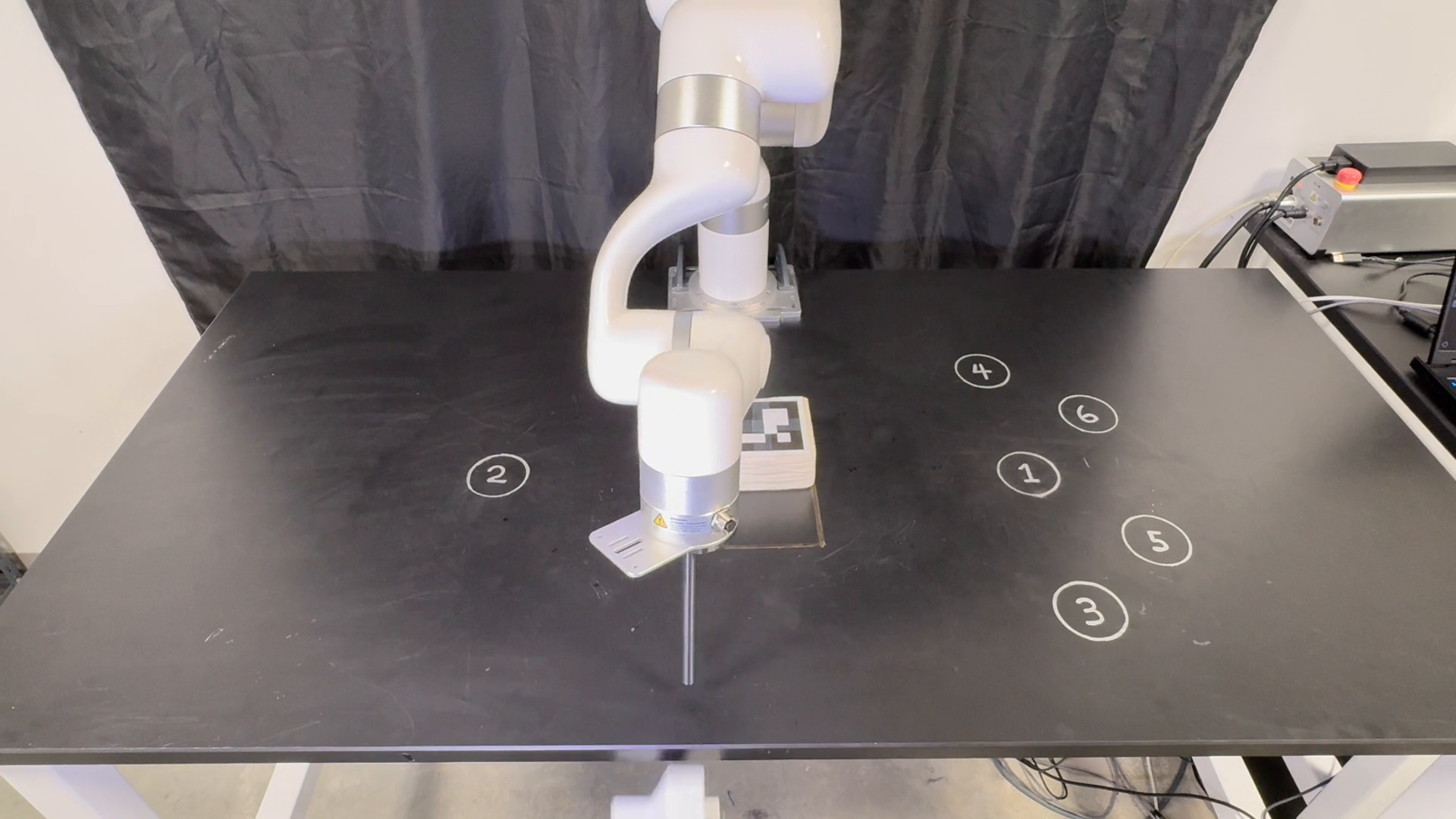} \\[-2pt]

        \rowlab{MPPI} &
        \includegraphics[width=0.16\linewidth]{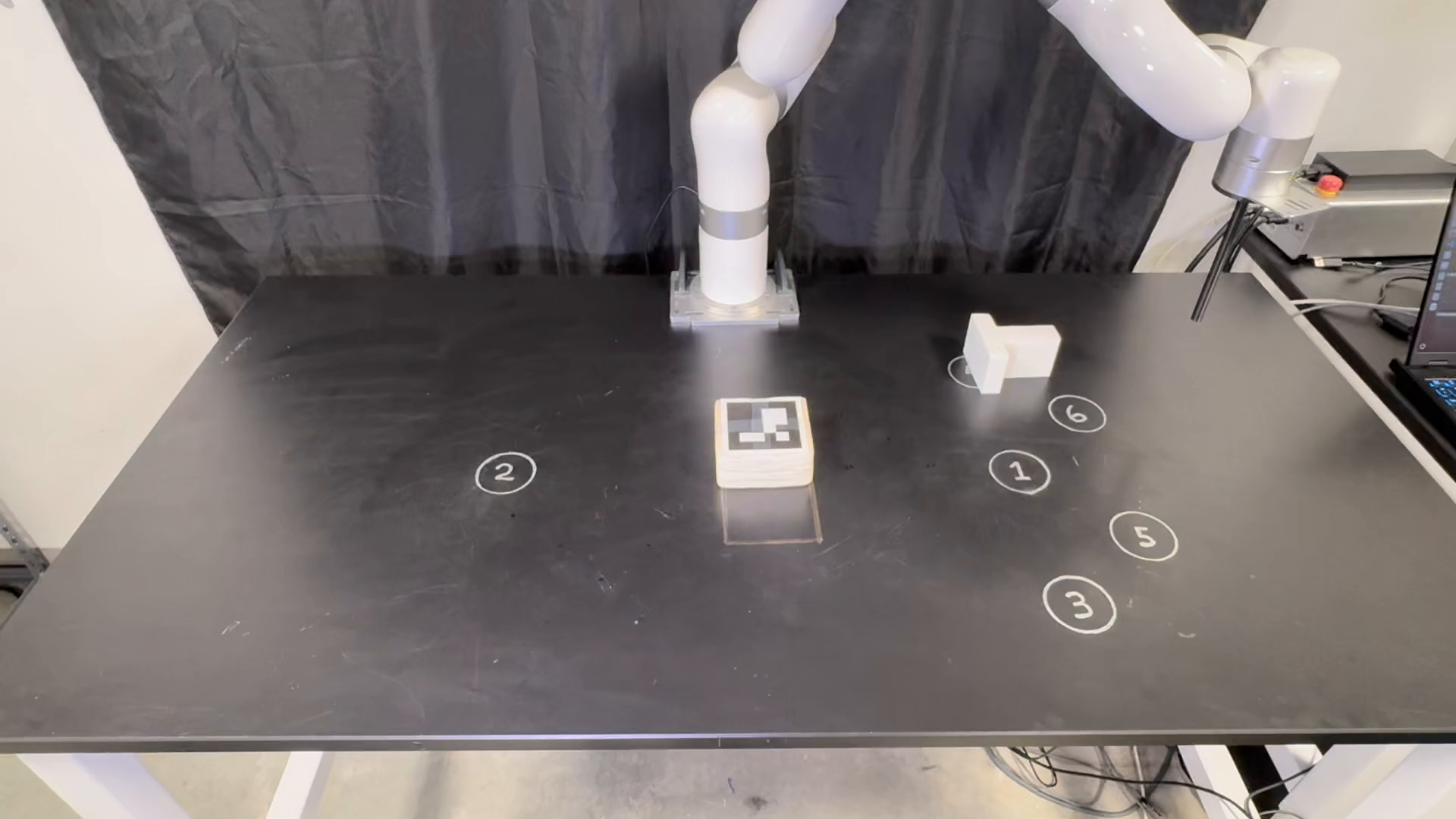} &
        \includegraphics[width=0.16\linewidth]{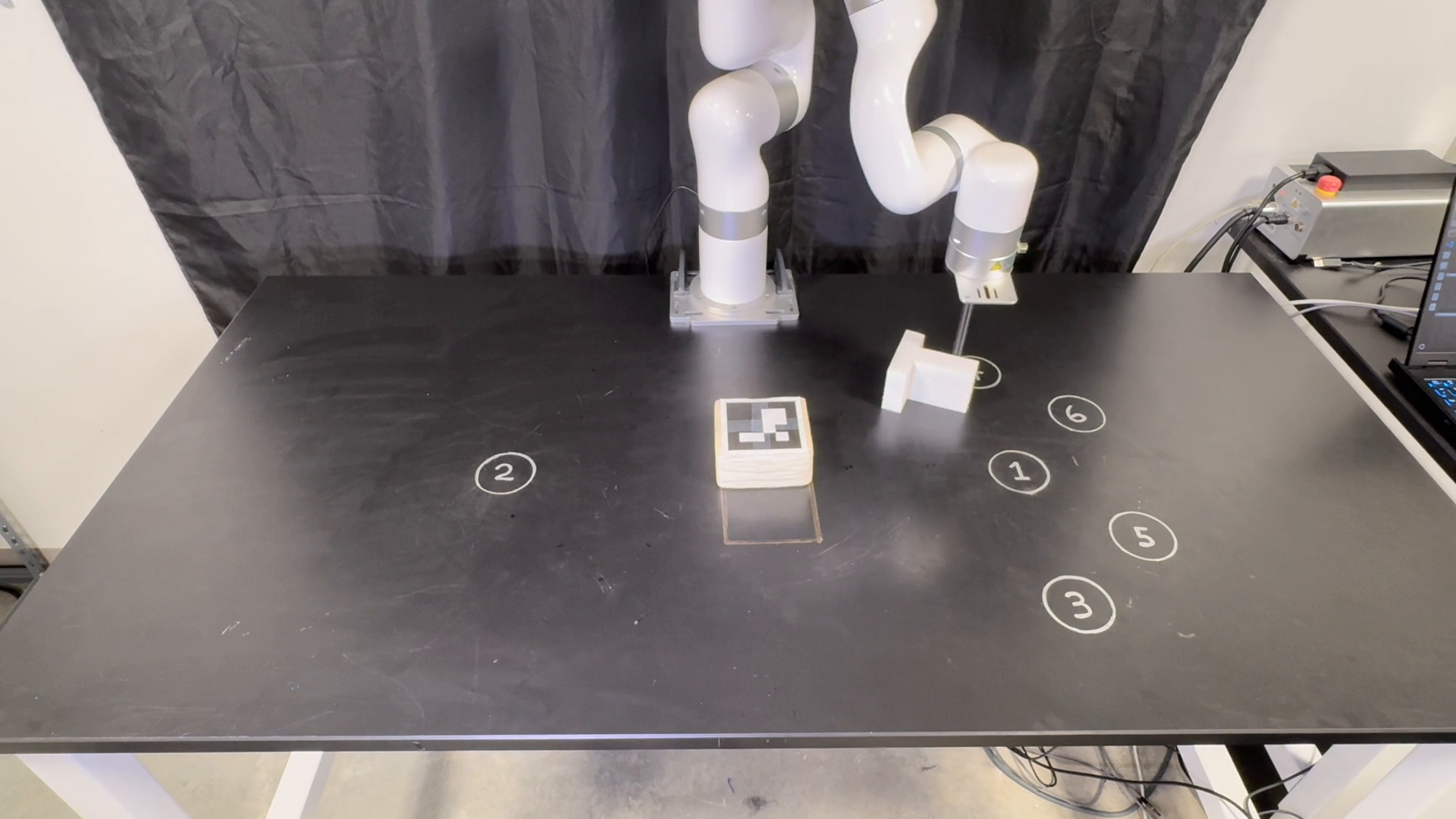} &
        \includegraphics[width=0.16\linewidth]{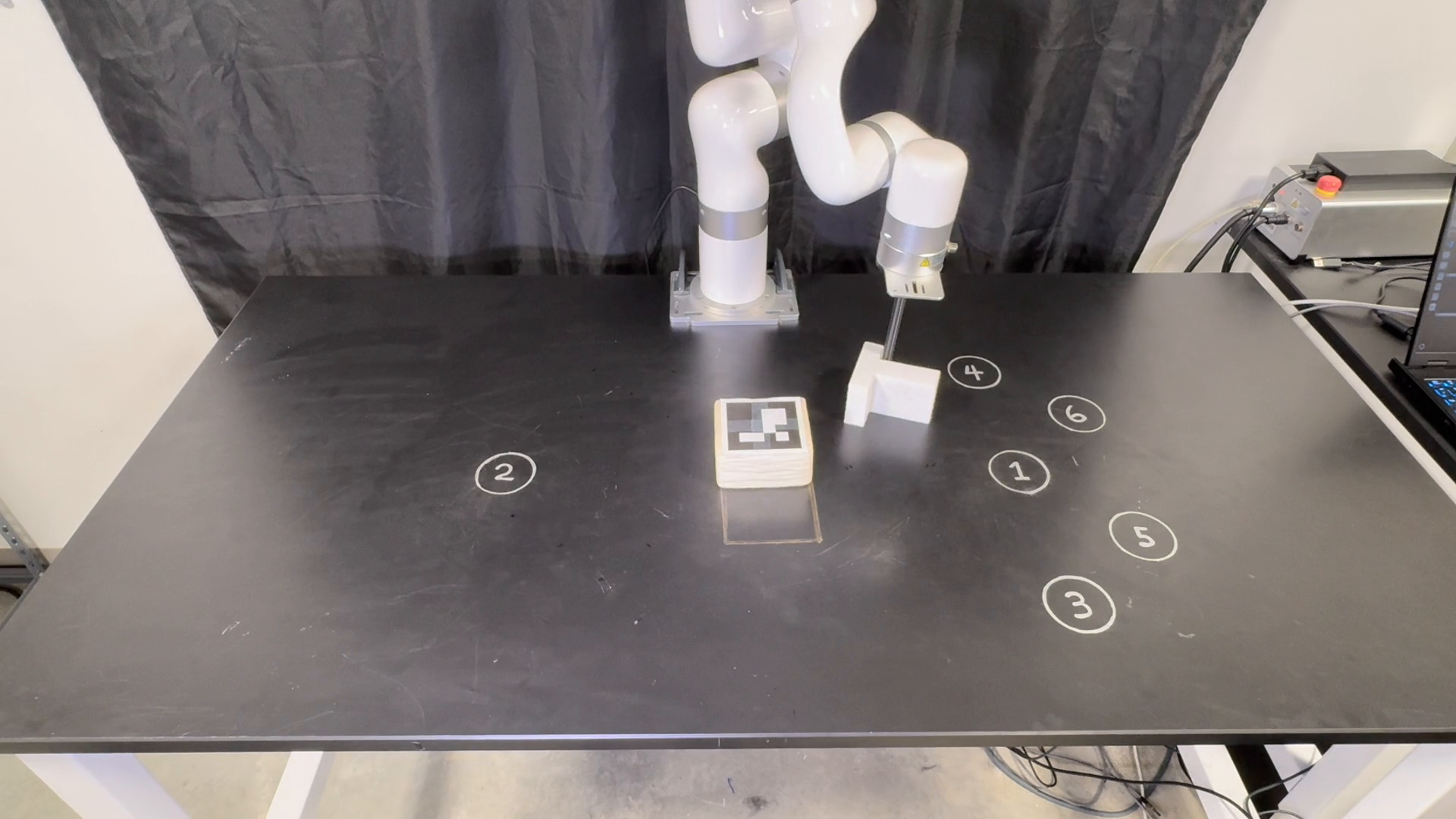} &
        \includegraphics[width=0.16\linewidth]{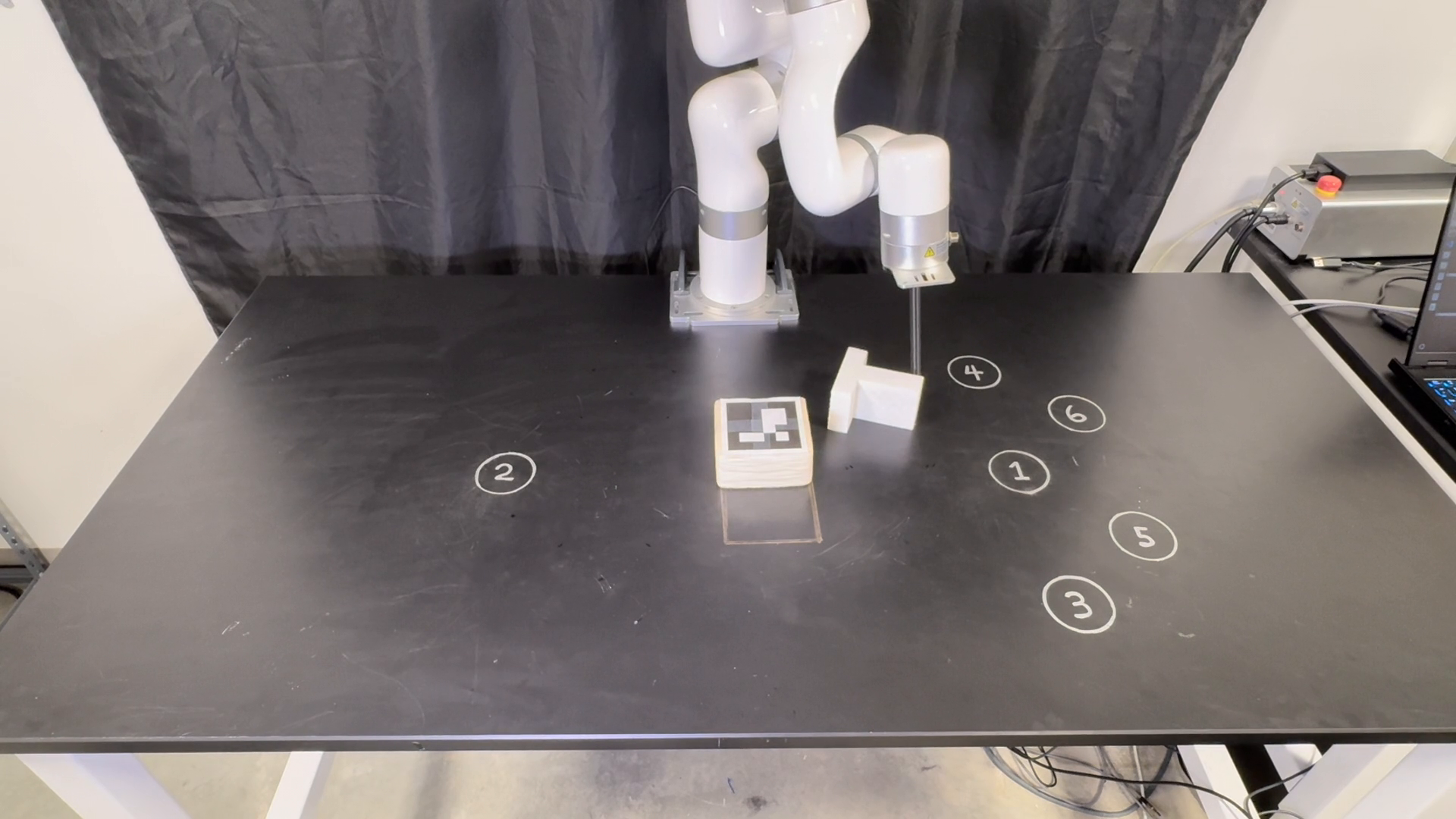} &
        \includegraphics[width=0.16\linewidth]{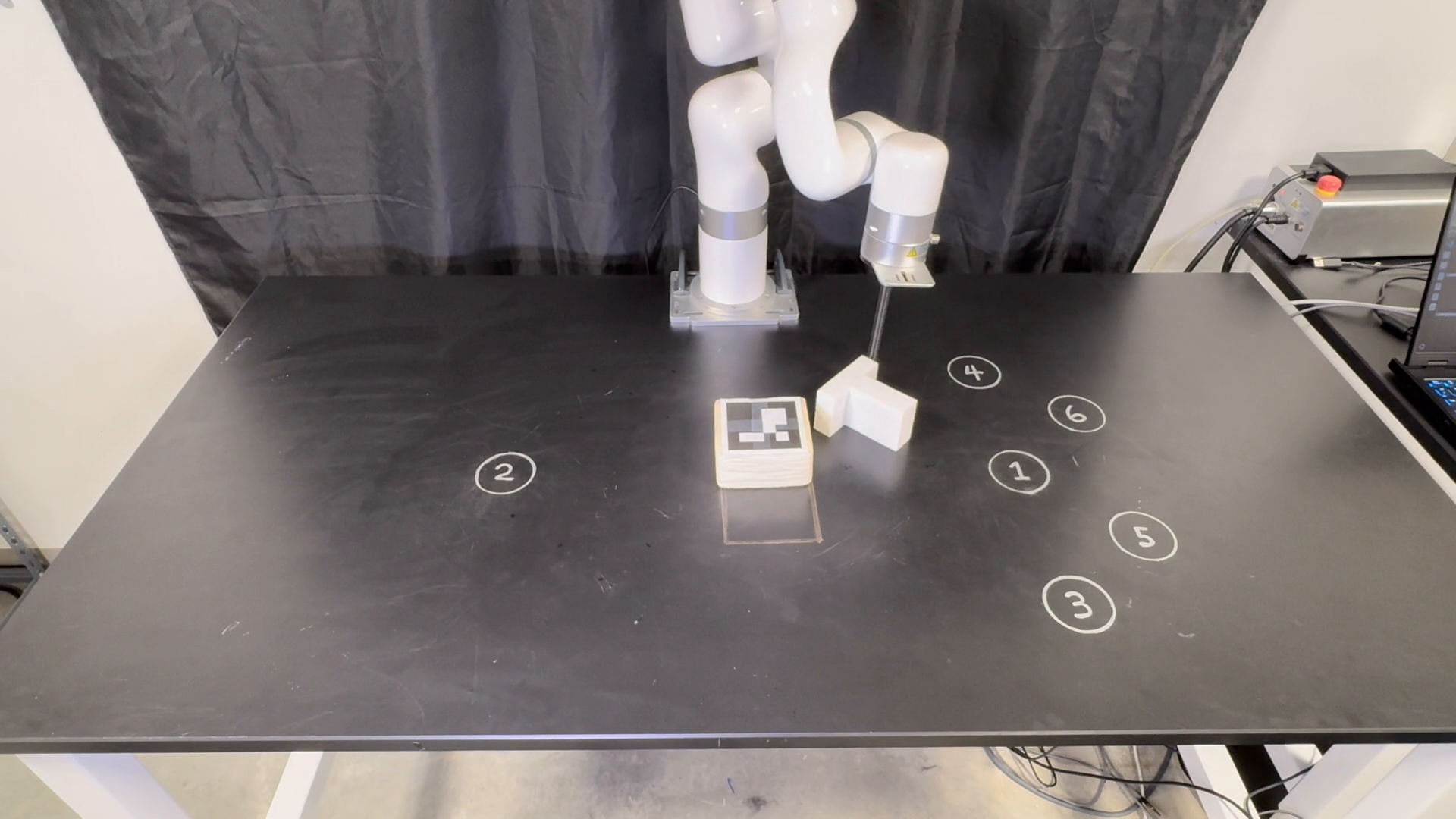} &
        \includegraphics[width=0.16\linewidth]{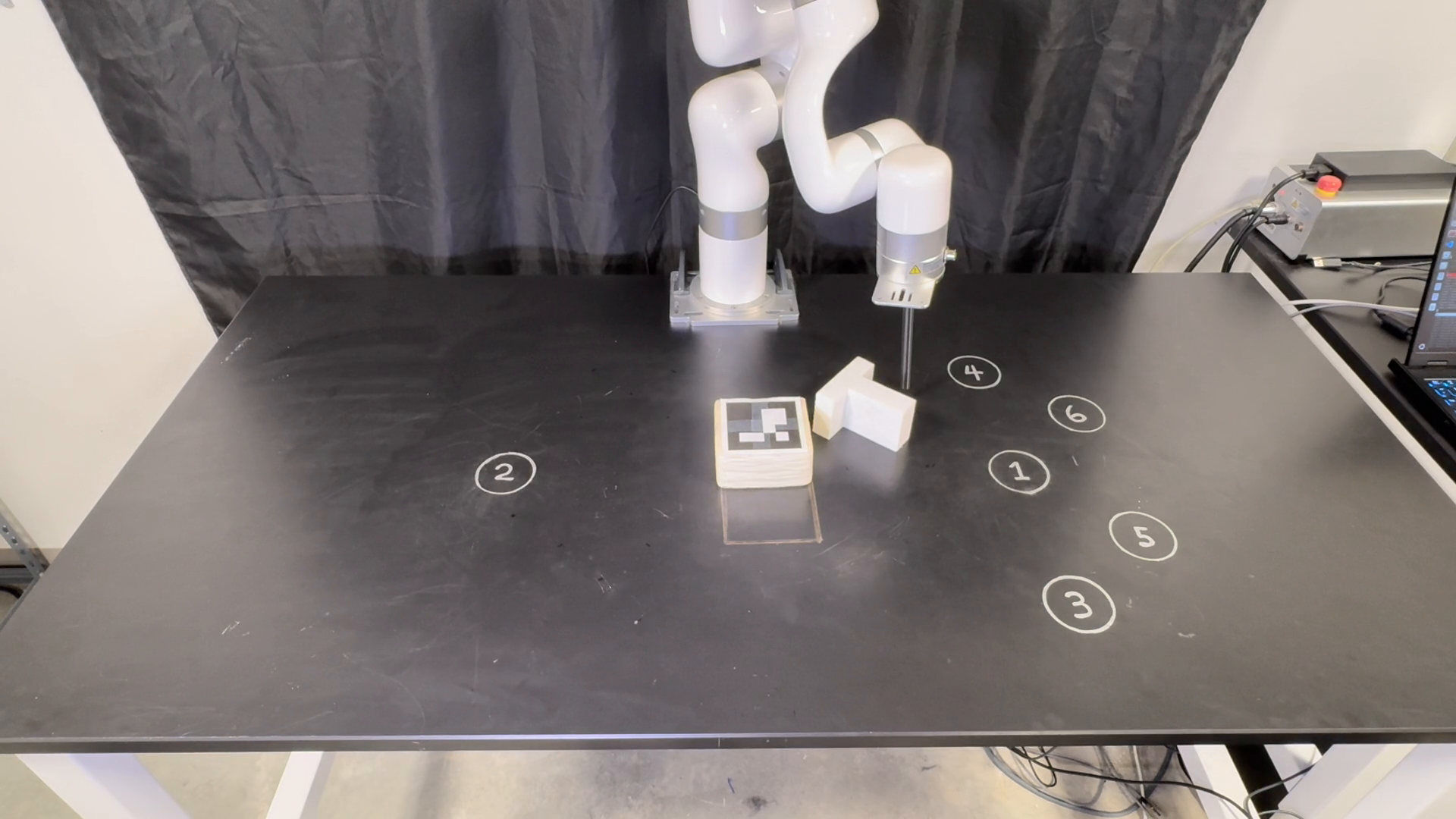} \\

    \end{tabular}
    \caption{\textbf{Hardware execution} represented as time series from left to right on the single-obstacle task, one row per method.}
    \label{fig:timeseries_real}
\end{figure*}

\begin{figure}[thpb]
    \centering
    \includegraphics[width=\linewidth]{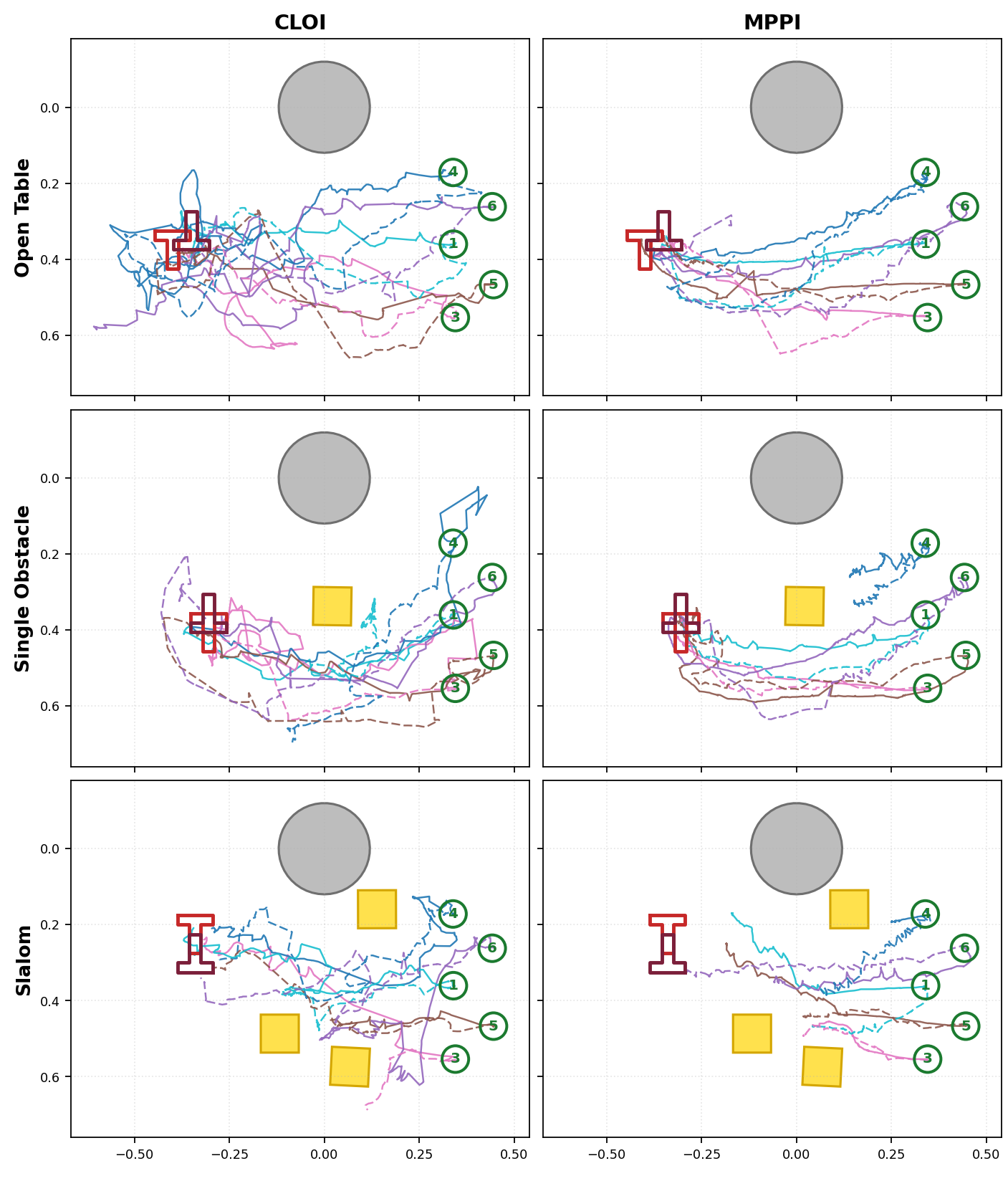}
    \caption{\textbf{Object trajectories on hardware.} Start positions are numbered in green, the robot in grey, and obstacles are shown in yellow; trajectories are colored by start position, solid for the $+90^\circ$ goal (red) and dashed for the $-90^\circ$ goal (dark red).}
    \label{fig:trajectories_real}
\end{figure}   

\begin{table}[t]
\centering
\caption{\textbf{Hardware experiments.} CLOI and MPPI on the xArm6, pushing a T-shaped block on the three tasks of Fig.~\ref{fig:trajectories_real}, open table, single obstacle and slalom. Each method is run for $10$ trials per task: five start positions, each pushed to $\pm90^\circ$ goals.}
\label{tab:real}
\setlength{\tabcolsep}{5pt}
\resizebox{\linewidth}{!}{\begin{tabular}{llccccccc}
\toprule
& & & \multicolumn{2}{c}{All trials} & \multicolumn{2}{c}{Successful} & & \\
\cmidrule(lr){4-5} \cmidrule(lr){6-7}
Task & Method & SR $\uparrow$ & $\epsilon_d$ & $\epsilon_\theta$ & $\epsilon_d^s$ & $\epsilon_\theta^s$ & $T$ $\downarrow$ & $\bar{f}$ $\uparrow$ \\
\midrule
\multirow{2}{*}{Open table}
  & MPPI & 0.50 & \textbf{0.363} & 0.916 & 0.427 & 0.863 & 14.3 & \textbf{8.9} \\
  & CLOI & \textbf{0.90} & 0.383 & \textbf{0.824} & \textbf{0.391} & \textbf{0.819} & \textbf{9.2} & 4.0 \\
\midrule
\multirow{2}{*}{Single obstacle}
  & MPPI & 0.40 & \textbf{0.321} & 0.990 & \textbf{0.386} & 1.099 & 17.0 & \textbf{9.0} \\
  & CLOI & \textbf{0.80} & 0.418 & \textbf{0.756} & 0.407 & \textbf{0.709} & \textbf{9.6} & 4.0 \\
\midrule
\multirow{2}{*}{Slalom}
  & MPPI & 0.00 & \textbf{0.449} & 1.355 & -- & -- & 20.0 & \textbf{10.9} \\
  & CLOI & \textbf{0.50} & 0.472 & \textbf{0.805} & \textbf{0.421} & \textbf{0.746} & \textbf{15.6} & 4.8 \\
\midrule
\multirow{2}{*}{\textbf{Mean}}
  & MPPI & 0.30 & \textbf{0.378} & 1.087 & 0.407 & 0.981 & 17.1 & \textbf{9.6} \\
  & CLOI & \textbf{0.73} & 0.424 & \textbf{0.795} & \textbf{0.406} & \textbf{0.758} & \textbf{11.5} & 4.3 \\
\bottomrule
\end{tabular}}
\end{table}
We evaluate CLOI in simulation and on hardware with a 6-DoF xArm6 manipulator. The task is for the robot to push a target object to a goal pose while avoiding static obstacles. 

\subsection{Experiment Setup}\label{subsec:experiment_setup}

\noindent \textbf{Metrics:} We report: \emph{success rate} (SR $\uparrow$), the fraction of trials whose final pose satisfies $\epsilon_d \le 0.05$\,m and $\epsilon_\theta \le 0.1$\, rad; trajectory-averaged \emph{position and orientation error}, over all trials ($\epsilon_d$, $\epsilon_\theta$ $\downarrow$) and over successful trials alone ($\epsilon_d^s$, $\epsilon_\theta^s$); \emph{execution time} ($T$ $\downarrow$), total run time; and \emph{planning frequency} ($\bar{f}$ $\uparrow$), the inverse mean wall-clock time per control step. For all experiments, a trial ends on reaching the goal tolerance or exceeding the maximum step count. 

\noindent \textbf{Baselines:}
We consider three baselines. \emph{MPPI}~\cite{pezzato2025samplingbasedmodelpredictivecontrol} applies sampling-based MPC monolithically to the joint robot--object state. \emph{MPPI (more iterations)} uses $8$ planning iterations, matching the compute budget of CLOI. This isolates the benefit of the consensus formulation from the benefit of more compute. \emph{C3+}~\cite{bui2025pushanythingsinglemultiobject} is contact-implicit MPC that also uses ADMM but splits along the complementarity constraints rather than the robot--object boundary.

\noindent\textbf{Cost Functions:}
We design costs $\ell_o$ and $\ell_r$ for object pushing on a support surface to a goal pose $g$ while avoiding obstacles. Let $d^2(\cdot,\cdot)$ denote distance in $SE(2)$ pose. Let $\phi_2$ be split into $\phi_{\mathrm{obs}}(b)$ and $\phi_{\mathrm{sup}}(b)$, the signed distances from a boundary point $b$ to the nearest obstacle and to the edge of the support surface. The object costs are
\begin{align}
    \ell_o(x_t^o, w_t^o) &= q_g\, d^2(x_t^o, g) + q_{\mathrm{sup}} \psi_{\mathrm{sup}}  + q_{\mathrm{obs}} \psi_{\mathrm{obs}} +  \notag \\
    &\quad q_{e}\|w_t^o\|^2  +q_{\Delta} \textstyle\sum_{t=1}^{H^{c-1}} \big\| w^o_t - w^o_{t-1} \big\|^2,
      \label{eq:obj_stage_cost}\\
    \ell_f(x_{H^c}^o) &= q_{g,f}\, d^2(x_{H^c}^o, g).\label{eq:obj_terminal_cost}
\end{align}
The obstacle term $\psi_{\mathrm{obs}}$ sums $e^{\left([\delta_{\mathrm{obs}} - \phi_{\mathrm{obs}}(b)]_+ / \eta\right)^2} - 1$ over obstacles $b$, exponentially penalizing obstacle proximity $\phi_{\mathrm{obs}}(b)$ with margin $\delta_{\mathrm{obs}}$. The support term $\psi_{\mathrm{sup}} = \big[ \phi_{\mathrm{sup}}(b) + \delta_{\mathrm{sup}} \big]_+^2$ quadratically penalizes leaving the table. 
The last two terms penalize large or abruptly changing wrenches. The robot cost encourages the end-effector to approach the object:
\begin{align}
    \ell_r(x_t^r, u_t^r) = \;& \sigma(x_t^o)\big(q_{a} \psi_{\mathrm{app}} + q_{\ell} \psi_{\mathrm{align}} + q_{v} \psi_{\mathrm{tilt}} + q_u\|u_t^r\|^2\big) \notag\\
    & + q_{h} \psi_{\mathrm{height}} + q_{c} \psi_{\mathrm{clear}}, \label{eq:robot_stage_cost}
\end{align}
The approach term $\psi_{\mathrm{app}} = \max(\|p_t^{ee} - p_t^{o}\|^2 - r_0^2, 0)$ draws the tip to within a radius $r_0$ of the object, and $\psi_{\mathrm{align}} = \max(\gamma_0 - \cos\angle(p_t^{o} - p_t^{ee},\, p_t^{o*} - p_t^{o}), 0)$ keeps it behind the local goal; $\psi_{\mathrm{tilt}}$, $\psi_{\mathrm{height}}$ and $\psi_{\mathrm{clear}}$ penalize end-effector roll and pitch, deviation from the z-plane, and robot--obstacle proximity. The fade $\sigma(x_t^o) \in [0,1]$ decays to zero at the goal so that posture shaping does not compete with the final approach, but is withheld from $\psi_{\mathrm{height}}$ and $\psi_{\mathrm{clear}}$, which enforce table and obstacle safety throughout the horizon.

\begin{table}[t]
\label{tab:cost_weight}
\centering
\caption{\textbf{Cost weights.} The weights used in \eqref{eq:obj_stage_cost}--\eqref{eq:robot_stage_cost} for every method and scenario within each setting. Goal entries show position/orientation and $q_\Delta$ for $[f_x, f_y, \tau]$. On hardware, $q_v$, $q_h$, and $q_c$ are enforced with an external safety controller. }
\label{tab:cost_weights}
\setlength{\tabcolsep}{3pt}
\small
\begin{tabular}{lcccccc}
\toprule
\emph{Object cost}: & $q_g$ & $q_{g,f}$ & $q_e$ & $q_\Delta$ & $q_{\mathrm{obs}}$ & $q_{\mathrm{sup}}$ \\
\midrule
Simulation & 200/16 & 200/16 & 0.01 & 8/8/2 & 1000 & 500 \\
Hardware   & 200/10 & 100/45 & 0.01 & 2/2/1 & 3000 & $5{\times}10^{5}$ \\
\midrule
\emph{Robot cost}: & $q_a$ & $q_\ell$ & $q_v$ & $q_u$ & $q_h$ & $q_c$ \\
\midrule
Simulation & 60 & 50 & 80 & 0.05 & 1.0 & 0.1 \\
Hardware   & 6000 & 600 & - & 0.05 & - & - \\
\bottomrule
\end{tabular}
\end{table}

\subsection{Simulation Results}
\label{subsec:simulation_experiments}

Simulated experiments run in MuJoCo MJX~\cite{todorov2012mujoco} on a single NVIDIA RTX 4090. We evaluate on six scenarios (Fig.~\ref{fig:tasks_sim}), using the same $25$ start--goal pairs, budget, and costs. The pushed object is a T-shaped block by default, but we also consider four YCB objects~\cite{calli2017ycbdataset} with identical settings in order to test CLOI's performance with different object geometry. 
We set parameters as $K = 256$, $H^c = 32$, $N_{\mathrm{ADMM}} = 4$, $\rho = 2.0$, $\nu = 0.8$, and $\lambda = 0.5$.

\noindent\textbf{Consensus objective:} From Table~\ref{tab:parameters}, object pose consensus has the highest SR among the consensus variables, followed by contact point and wrench ($0.96$, $0.95$, $0.93$). Wrench consensus performs worst because the two blocks realize it differently: the object block commands a wrench directly, whereas the robot produces it through contact, so a small change in desired wrench can require a long end-effector motion. The proposed object pose consensus is invariant to contact modes and is therefore the most stable quantity to agree on.

\noindent\textbf{Parameter sensitivity.} From Table~\ref{tab:parameters}, we further observe that SR rises with sample count ($0.87 \to 0.96$ from $K = 64$ to $512$), as broader exploration allows finding better actions, and with the number of ADMM rounds ($0.80 \to 0.95$), which tighten the agreement between the blocks. Nonetheless, the planning frequency falls from $3.6$ to $0.6$\, Hz. A smaller consensus penalty is preferable ($0.94$ at $\rho \le 2.0$ against $0.84$ at $\rho = 8.0$): a large $\rho$ forces each block to consensus prior to converging on low task-cost solutions. Leaning the consensus toward the object block is advantageous ($0.94$ at $\nu = 0.8$ versus $0.74$ at $\nu = 0.2$), because the object subproblem converges quicker and has a more reliable estimate of the solution. A low temperature is preferred ($0.96$ at $\lambda = 0.1$), reserving robot-level exploration for finer corrections and leaving the object subproblem to provide a higher-level direction. Overall, every variation shows the object-level planner is beneficial.

\noindent\textbf{Comparison against baselines.} CLOI outperforms both MPPI baselines on every scenario of Table~\ref{tab:ablation}. Granting standard MPPI a matched compute budget with $8$ additional planning steps only increases the success rate by $0.04$, which shows that CLOI's performance comes from the decomposition rather than additional compute. We observed that the gap is the widest in scenarios that require reorientation of the object for obstacle avoidance, such as the sign and slalom scenarios. This is because the object subproblem plans the reorientation explicitly instead of discovering it through sampling. On all trials, CLOI has lower position and orientation errors ($0.470$, $1.233$) and a shorter execution time ($33.3$ against $43.5$), reflecting the frequent stalls the baselines experience near an obstacle. On successful trials, MPPI attains a marginally lower position error ($0.447$ against $0.450$), since the trials it completes are the short, direct pushes that require minimum reorientation. The limitation is planning frequency, $1.1$\, Hz (CLOI) against $9.5$\, Hz (MPPI).

C3+ reaches an SR of $0.66$ on the open table but $0.21$ when averaged over all scenarios, as splitting along the complementarity constraints resolves contact locally but lacks an object-level plan to follow around obstacles. Therefore, it is frequently stuck. In successful trials, C3+ is the most precise with the lowest position and orientation errors ($\epsilon_d^s = 0.348$, $\epsilon_\theta^s = 0.971$). However, it shows a long execution time ($267.2$) due to frequent mode switches, where the end-effector lifts up and repositions around the object. This led to timeout failures and hence low SR in the open table.

\subsection{Hardware Experiments}
\label{subsec:hardware_experiments}

We deploy CLOI on a real xArm6 manipulator using NVIDIA RTX 3080Ti. Object pose is estimated by FoundationPose~\cite{wen2024foundationposeunified6dpose} from an Intel RealSense D-455, and a ROS\,2 interface relays the measured states to the planner and the joint velocities to the arm at a 50$\mathrm{ms}$ control period. We evaluate the three scenarios of Fig.~\ref{fig:trajectories_real}: open table, single obstacle, and slalom. Fig.~\ref{fig:timeseries_real} shows representative successful and failed rollouts. CLOI uses $K = 256$, $H = 24$, $N_{\mathrm{ADMM}} = 2$, $\rho = 2.0$, $\nu = 0.8$, and $\lambda = 10$, while MPPI uses $K = 256$, $H = 32$, and $\lambda = 10$. 

\noindent\textbf{Open table.} In Table~\ref{tab:real}, CLOI shows higher SR ($0.90$) compared to MPPI ($0.50$). CLOI outperforms MPPI even without obstacles because it plans the object's reorientation explicitly as opposed to random sampling. Nearly all of MPPI's failures are in orientation: although the position error is the lowest ($0.363$), it exhausts the time budget before the final reorientation, evidenced by longer execution time ($14.3$ against $9.2$). 
MPPI's successes are mostly near-straight pushes, which leads to low position error. CLOI shows lower orientation error ($0.824$ against $0.916$). Its only failure occurs when the planned reorientation moves the object out of the arm's reachable workspace, as shown in Fig.~\ref{fig:timeseries_real}.

\noindent\textbf{Single obstacle.} CLOI maintains a SR of $0.80$ with the lowest orientation error of either scenario ($0.756$). Its two failures are similar to the open-table: the object leaves the reachable workspace during reorientation, as shown in Fig.~\ref{fig:timeseries_real}. Meanwhile, the SR of MPPI drops to $0.40$, failing either by stalling in front of the obstacle, recording the longest execution time ($17.0$s), or running out of time during reorientation, where it attains the worst orientation error ($1.099$). The SR remains comparable to the open table because it still succeeds in start-goal pairs that admit a near-straight push, but not in those requiring a detour. 

\noindent\textbf{Slalom.} The gap is the widest because the scenario requires a complex detour and a reorientation. CLOI succeeds in 50\% of the trials while MPPI succeeds in none. This is because CLOI is guided toward the required object motions, whereas MPPI must discover these through random sampling.

\noindent\textbf{Transfer from simulation.} The ranking and trends are consistent. CLOI increases the mean SR from $0.30$ to $0.73$, lowers orientation error ($0.795$ against $1.087$), and reduces execution time by $33\%$ ($11.5$ against $17.1$). CLOI generally shows lower errors in $\epsilon_\theta$, $\epsilon_d^s$, and $\epsilon_\theta^s$. One exception is the all-trials position error, where CLOI is higher ($0.400$ against $0.342$). This is because CLOI spends more time reorienting the object to the goal, whereas MPPI makes direct pushes to the goal position but fails to reorient, as depicted in Fig.~\ref{fig:trajectories_real}. Planning frequency is higher than in simulation ($4.3$ against $1.1$\, Hz) because hardware uses $N_{\mathrm{ADMM}} = 2$ rather than $4$, trading consensus rounds for rate.

\section{Conclusion}
\label{sec:conclusion}

We presented CLOI, a sampling-based MPC method that guides high-dimensional non-prehensile manipulation planning with low-dimensional object-only planning, exploiting the physical structure. The two are coordinated through agreement on the object pose trajectory using the ADMM algorithm. Across six simulated scenarios and five objects, CLOI increases the success rate from $0.29$ to $0.64$ compared to standard MPPI on the joint robot-object problem at a matched computational budget, and from $0.30$ to $0.73$ on an xArm6 robot. Two limitations remain: each consensus round costs one update per block, increasing the planning time, and the object-level planner ignores the robot's reachable workspace. Future work will incorporate reachability into the object-level planner and randomize object friction and pose during rollouts for sim-to-real robustness.


\section*{Acknowledgment}
Conceptualization: NR, BL, NA; Methodology: NR, HK, SM, BL, NA; Theoretical analysis: NR, HK, SM; Algorithm development: NR, HK, SM, BR, CY, BL; Simulation validation and hardware evaluation: NR, HK, SM, BR, CY; Writing -- initial draft: NR, SM, BR, CY, BL; Writing -- review and editing: all authors; Supervision: NA.
\bibliography{bib/bib_string.bib,references.bib}

\addtolength{\textheight}{-12cm}   

\end{document}